\documentclass{ieeeaccess}
\usepackage{cite}
\usepackage{amsmath,amssymb,amsfonts}
\usepackage{graphicx}
\usepackage{textcomp}
\usepackage{caption}
\usepackage{subcaption}
\usepackage{gensymb}
\usepackage{nccmath}
\usepackage{amssymb}
\usepackage{float}
\usepackage{enumitem}
\usepackage{colortbl}
\usepackage{makecell}
\usepackage{pbox}

\usepackage{algorithmic}

\usepackage[usestackEOL]{stackengine}

\usepackage[ruled,vlined]{algorithm2e}
\def\BibTeX{{\rm B\kern-.05em{\sc i\kern-.025em b}\kern-.08em
    T\kern-.1667em\lower.7ex\hbox{E}\kern-.125emX}}

\DeclareMathOperator*{\argmin}{arg\,min}

\definecolor{Gray}{gray}{0.85}
\newcolumntype{a}{>{\columncolor{Gray}}c}

\usepackage{epstopdf}
\AppendGraphicsExtensions{.tif}

\begin{document}
\strutlongstacks{T}
\makeatletter
\algocf@newcmdside@kobe{Init@Initialization}{%
    \KwSty{Initialization:}%
    \ifArgumentEmpty{#1}\relax{ #1}%
    \algocf@group{#2}%
    \par
}

\newcommand\Init[1]{%
    \Init@Initialization{#1}%
}
\makeatother
\makeatletter
\algocf@newcmdside@kobe{Alg@Algorithm}{%
    \KwSty{Algorithm:}%
    \ifArgumentEmpty{#1}\relax{ #1}%
    \algocf@group{#2}%
    \par
}

\newcommand\Alg[1]{%
    \Alg@Algorithm{#1}%
}

\history{Date of publication xxxx 00, 0000, date of current version xxxx 00, 0000.}
\doi{xx.xxxx/ACCESS.2020.DOI}

\title{Unsupervised Doppler Radar Based Activity Recognition for e-Healthcare}
\author{\uppercase{Yordanka Karayaneva}\authorrefmark{1}, 
\uppercase{Sara Sharifzadeh}\authorrefmark{1}, \uppercase{Wenda Li}\authorrefmark{2},
\uppercase{Yanguo Jing}\authorrefmark{1},
\uppercase{Bo Tan}\authorrefmark{3}
}
\address[1]{Faculty of Engineering, Environment and Computing, Coventry University, UK (e-mail: (karayany,ac8115,ac2716)@coventry.ac.uk)}
\address[2]{University College London,UK (email:wenda.li@ucl.ac.uk)}
\address[3]{Tampere University, Finland (email: bo.tan@tuni.fi)}

\tfootnote{} 

\markboth
{Karayaneva \headeretal: Preparation of Papers for IEEE TRANSACTIONS and JOURNALS}
{Karayaneva \headeretal: Preparation of Papers for IEEE TRANSACTIONS and JOURNALS}

\corresp{Corresponding author: Bo Tan (e-mail: bo.tan@tuni.fi).}

\begin{abstract}
Passive radio frequency (RF) sensing and monitoring of human daily activities in elderly care homes is an emerging topic. Micro-Doppler radars are an appealing solution considering their non-intrusiveness, deep penetration, and high-distance range. Unsupervised activity recognition using Doppler radar data has not received attention, in spite of its importance in case of unlabelled or poorly labelled activities in real scenarios. This study proposes two unsupervised feature extraction methods for the purpose of human activity monitoring using Doppler-streams. These include a local Discrete Cosine Transform (DCT)-based feature extraction method and a local entropy-based feature extraction method. In addition, a novel application of Convolutional Variational Autoencoder (CVAE) feature extraction is employed for the first time for Doppler radar data. The three feature extraction architectures are compared with the previously used Convolutional Autoencoder (CAE) and linear feature extraction based on Principal Component Analysis (PCA) and 2DPCA. Unsupervised clustering is performed using K-Means and K-Medoids. The results show the superiority of DCT-based method, entropy-based method, and CVAE features compared to CAE, PCA, and 2DPCA, with more than 5\%-20\% average accuracy. In regards to computation time, the two proposed methods are noticeably much faster than the existing CVAE. Furthermore, for high-dimensional data visualisation, three manifold learning techniques are considered. The methods are compared for the projection of raw data as well as the encoded CVAE features. All three methods show an improved visualisation ability when applied to the encoded CVAE features.
\end{abstract}


\titlepgskip=-15pt

\maketitle

\section{Introduction} \label{SecI:Introduction}
\PARstart{H}{uman} activity recognition for smart healthcare is an emerging topic. It is becoming even more prominent with the complications of ageing population worldwide. The population aged 65+ in the UK was 11.8 million in 2016, while this number is projected to grow to 20.4 million by 2041 \cite{ONS:Report}. Chronic and long-term conditions are well-known to increase with age. It is reported that 29\% of those aged 60-64 had a chronic condition, while the percentage grows to 50\% for elderly populations aged 75 or over. The implications of ageing with chronic conditions prevent elderly people from independent living. Thus, they are dependent on social care services such as living in care homes.

The demand for human activity detection and monitoring has rapidly increased over the past years. A number of devices are proposed including cameras, wearable technologies, infrared sensors, and radars. These devices are expected to provide daily monitoring of elderly people's activities and vital signs. Hence, this will provide peace of mind for their relatives regarding the physical health and mental health of the care home residents.
Cameras are often seen as an obvious and traditional solution for capturing observable data including human activities for subsequent recognition \cite{Cameras:HAR}, \cite{MobileCamera} \cite{Cameras:HAR2}. Video-depth cameras are capable of obtaining extremely high-resolution data, which can contribute to the detailed analysis of daily human activities. Nevertheless, camera devices suffer from intrusiveness, which is highly undesirable in the contexts of residential environment. The modern healthcare is concerned with the privacy and dignity of patients. Therefore, vision-based solutions are not recommended in smart care homes. 

Wearable sensor technologies are an effective solution for smart healthcare applications as they provide a combination of human activity recognition and vital signs detection \cite{Wearables:Review}, \cite{Wearables:Healthcare}, \cite{Wearables:Healthcare2}, \cite{Wearables:HealthcareReview}. Wearable sensors have the ability to capture small fractions of the body such as the movement of fingers \cite{Wearables:Overview}. Additionally, wearable sensing technologies can detect physiological signals such as heart rate and speech patterns \cite{Wearables:Review2}. However, the disadvantages and challenges of this technology are not to be under-rated. Wearable sensors are known to have poor battery life \cite{Wearables:Battery}. As they are "wearable", elderly populations may easily forget to wear the device or feel uncomfortable wearing it \cite{Wearables:Drawbacks}. 

Infrared sensors utilize human's body temperature distinguished from the lower ambient temperature in order to capture and detect human activities. Most IR sensors obtain ultra low-resolution data, where a subject identification is avoided. Thus, IR devices represent an attractive solution to be deployed in care homes and hospitals. Current studies reveal significant recognition rates (>90\%) for activities including standing, sitting, walking, falling, and others \cite{IR:PreviousWork}, \cite{IR:Paper}, \cite{IR:Paper2}. Contrarily to their advantages, IR devices suffer from a relatively low detection distance. It has been shown in a recent paper \cite{Karayaneva:IEEESensors} that the performance drops with distance growth, although not significantly. As IR sensors are low-resolution capturing devices, they lack sensitivity towards small fractions of the human body, which prevents more specific activities detection.


Passive Micro-Doppler radars are an appealing solution for human activity recognition. That is due to their non-intrusiveness, high distance range, deep penetration, and reliable accuracy rates \cite{Doppler:Activities}, \cite{Doppler:Activities2} \cite{Doppler:Activities3}. In addition, the passive radar uses the existing radio bursts in the environment. It avoids to bring extra RF source to aggravate the increasing electromagnetic interference in the residential environment. While passive Micro-Doppler radars traditionally have applications in human activity recognition \cite{Doppler:Activities}, they have also been deployed for vitals sign monitoring such as respiration \cite{Doppler:Healthcare}. In addition to their applications, the devices have been used for gait patterns analysis \cite{Doppler:Gait}.




Micro-Doppler radars have been extensively used for activity recognition with a focus on healthcare purposes \cite{Doppler:Healthcare}, \cite{Doppler:Pilot}, \cite{Doppler:Healthcare2}. Currently, majority of studies are based on pipelines, which are totally supervised or consist of a combination of unsupervised and supervised approaches. In most cases, the pipelines are based on unsupervised feature extraction methods, such as conventional PCA and Singular Value Decomposition (SVD) techniques. That is usually followed by a supervised classification method, such as Support Vector Machine (SVM) and k-Nearest Neighbours (k-NN) \cite{SVM:Doppler}, \cite{Previous:SVM}. In the pursuit of a more accurately measured covariance matrix from PCA, variations of PCA have been used for Micro-Doppler data. In \cite{PCA:L1}, the authors applied L1 norm PCA opposed to standard PCA and achieved improved testing accuracies. Furthermore, 2DPCA has been compared with standard PCA for Doppler radar data \cite{Compare:PCA2DPCA}. Considering the fact that 2DPCA accepts 2D image matrices as an input, the dependencies of the pixels are retained. The results of that work revealed improved recognition rates for 2DPCA by more than 10\%. Furthermore, unsupervised PCA has been combined with supervised Linear Discriminant Analysis (LDA) and shallow neural networks (SNN) \cite{3DPCA}. The proposed architecture in that work was the first to use a 3D-signal representation by retaining the matrix dependencies. Results reveal better performance than conventional PCA and 2DPCA for Doppler radar data. 

In a pilot study, the \textit{Doppler-Radar-2018} dataset was used. The work employed Hidden Markov Models (HMM) in order to extract activity information from each Doppler sequence \cite{Doppler:Pilot}. The output of the HMM training was clustered using K-Means and K-Medoids. The Kullback-Leibler (KL) log-likelihood with K-Medoids for clustering obtained the highest accuracy. HMM is a supervised framework that requires the labels and generates  log-likelihood values as a measure of similarity of a candidate sample to each of the classes. Therefore, log-likelihood values can be used for decision making directly and the idea of using them as a feature and applying unsupervised methods such as K-Means for clustering them is not the best analysis pipeline. HMM was also used in another previous work \cite{Wenda:PreviousWork}, for classification of extracted physical features where 72\% accuracy was achieved. 

Another group of supervised techniques are based on deep learning approaches. These methods require more data for learning their objective functions. Recently, Convolutional Neural Networks (CNNs) and Recurrent Neural Networks (RNNs) architectures have been used. Furthermore, CAE was used for feature extraction for Doppler radar data \cite{Radar:Activity}. That was followed by classification based on a supervised framework by fine-tuning and a Softmax classifier. In these techniques, feature extraction from data streams are performed automatically with minimum user required settings \cite{NNs:Features}.

Unsupervised learning is yet a minimally researched topic for Doppler radar based applications. The advantage of unsupervised methods compared to supervised techniques is that they do not require labeling data. This usually influences the accuracy of unsupervised methods compared to supervised techniques. That is because in the absence of labels, the learning is only guided based on the input variables, their variations and characteristics. That does not necessarily help to learn the decision rules correctly. However, the models can be updated faster compared to the supervised strategies. Hence, the learning capacity of the latter are limited due to labeling requirement for any new coming data. In practical settings, usually recognition of few activities is critical such as a fall or immobility. In future, such activities can be labelled and recognized among the clustered activities. However, this work is only focused on unsupervised activity clustering of Doppler radar data and the latter problem is not addressed in this paper. It will be considered for future studies. In addition, unsupervised learning usually requires the use of techniques for estimating the number of clusters. This is due to the fact that subjects conduct a broad number of activities in a real world scenario. As such, embedding all activities in a pre-collected dataset for supervised frameworks is problematic.

Unsupervised learning methods can be categorised into two groups of manual or automated feature extraction strategies. In rule-based systems, specially those strategies based on hand-crafted features, prior knowledge about the experimental system and environment are important \cite{Handcrafted}. Besides that, careful selection of the feature extraction technique and manual selection of some parameters depending on the setup rules are required. For example, the signal strength and angle of measurement can influence filtering window size and scaling or choice of basis function in spatio-temporal feature extraction techniques. On the other hand, in automated feature extraction approaches such as CAEs, such prior settings are not required. The embedded objective function and optimization removes the need for manual settings \cite{NNs:FE}, \cite{NNs:FE2}. Nevertheless, the computational time for automated feature extraction approaches is more expensive compared to manual feature extraction.

In this paper, the \textit{Doppler-Radar-2018} dataset, that was used previously in \cite{Doppler:Pilot}, is considered. An unsupervised framework is developed despite labels availability. As explained earlier, the importance of the designed framework is the applicability to projects with poor labeling scenarios. For this aim, four groups of unsupervised feature extraction strategies are considered: (1) frequency-domain analysis based on 2D DCT (2) entropy analysis (3) convolutional filtering strategies based on CVAE and CAE (4) unsupervised PCA analysis including 1D and 2D analysis. For DCT and entropy feature extraction methods, two methods are proposed. The extracted features are clustered into different activity groups based on unsupervised clustering strategies using K-Means and K-Medoids. In order to evaluate the results, the known labels are utilised only at the result evaluation step. Leave-one-subject-out cross validation (LOOCV) is used so that, the built models are tested on unseen data of one subject. Due to the fact that in unlabelled conditions the number of classes is unknown, four unsupervised metrics, namely Elbow, Silhouette, Davies-Bouldin and Dunn's index are used.



The contributions of the study to the research community are the following:
\begin{itemize}
    \item \textbf{Two proposed unsupervised feature extraction methods for Doppler radar data:} The Doppler radar data in this study has high dimensionality. When reshaped into 2D maps, there are different distinguished patterns for each activity. On the other hand, the high dimensionality leads to an ill-posed problem and over-fitting. Therefore, feature extraction from 2D image maps is employed. For this aim, the local areas with low level of variation and insignificant information can be cancelled out from the analysis. In order to extract the most meaningful information, two local patching and feature extraction methods are proposed in this paper. To retain the unsupervised scenario and evaluate the features, the Dunn's index is used. It is used as a criterion for evaluation of the activity clustering results, using the locally extracted features. That also helped to choose the first proposed feature extraction method's parameters such as the local patch size and location. The first proposed method uses 2D DCT to extract features from local patches of the 2D images. The second proposed method is based on entropy of the local patches. The average testing results showcase 5\%-10\% improvement by using the proposed techniques for different scenarios compared to the previous methods. To the best of our knowledge, such unsupervised algorithms have not been used for human activity recognition using Doppler radar data.
    
    \item \textbf{Comprehensive study of unsupervised learning for Doppler radar data:} This work is a pioneering study concerning unsupervised learning for Doppler radar data. Based on a comprehensive study, four different metrics are used to estimate the number of clusters in the unsupervised framework. That is useful in real scenarios, where the number of activities can be high and recognizing few of them among all clustered activities is required. Additionally in this paper, four groups of unsupervsed feature extraction strategies are compared: the proposed methods based on (1) local 2D DCT and (2) local entropy are compared by (3) architectures using deep CVAE and CAE, where the former has not been used previously for Doppler radar data, and (4) previous methods using PCA and 2DPCA. The extracted features are clustered with K-Means and K-Medoids. The proposed methods based on local DCT and local entropy, and CVAE achieved around 5\%-20\% higher average testing accuracy in comparison with CAE, PCA, and 2DPCA. The proposed methods for feature extraction along with CVAE encoded features can be useful for unsupervised cases or semi-supervised cases with poor labeling.
    
    \item \textbf{High-dimensional data visualisation enhancement:} Manifold learning methods for high-dimensional data visualisation are considered in this study. Doppler radar dataset can benefit in regards to activities data visualisation. This can reveal similarities between certain activity groups. The manifold learning methods are known to provide good separation between the classes \cite{VisualMethods}. In this study, CVAE encoded data have been used for data visualisation improvement. For the first time for Doppler radar data, in this research the manifold learning methods' performance over raw data and encoded data using CVAE is employed to illustrate any improvement for the sake of visualisation. The three methods for high-dimensional data visualisation t-Distributed Stochastic Neighbour Embedding (t-SNE), Multidimensional Scaling (MDS) and Locally Linear Embedding (LLE) are compared in the two scenarios. Initially, the methods are used to transform the raw data $D_{n \times 6400}$ to a 2-dimensional space. Secondly, the transformation is performed on the encoded features by CVAE data $D_{n \times 50}$. Comparison of the results reveal better separations of the clusters using the three methods, when CVAE encoding is used. This showcases the strength of CVAE for data separation. Hence, CVAE encoded features can be used in manifold learning for visualisation purposes.
\end{itemize}
The rest of the paper is organized as follows: In Section \ref{SecIII:Methodology} the methods including the database description, number of clusters estimation and the proposed approaches for feature extraction and clustering are described. Additionally, three methods for data visualisation are defined. The results for unsupervised feature extraction and clustering are shown in Section \ref{SecIV:Results}. Then, data visualisation techniques are compared by transforming the raw data as well as the CVAE encoded data. Section \ref{Section:Discussion} critically evaluates the main findings of the study, including the proposed architectures for unsupervised learning and the manifold learning methods for high-dimensional data visualisation. Finally, Section \ref{Section:Conclusion} concludes the study with the most valuable outcomes.

\section{Methodology} \label{SecIII:Methodology}
\begin{figure}[h]
\centering
\includegraphics[scale=0.43]{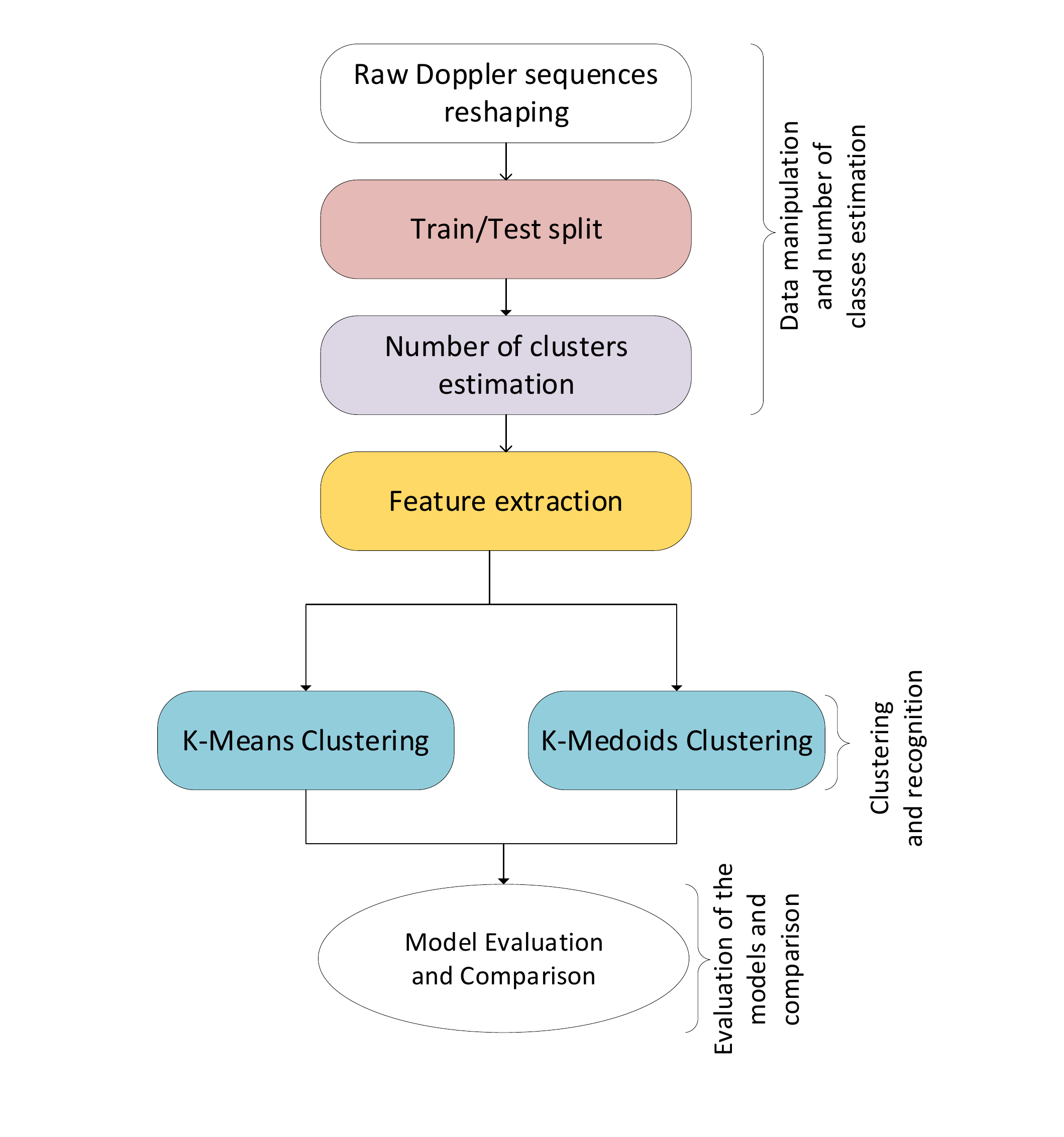}
\caption{Flowchart, describing the overall analysis framework of the paper. }
\label{fig:flowchart}
\end{figure}

The unsupervised framework in this study consists of a number of steps as illustrated in Fig. \ref{fig:flowchart}. The first step is to divide data into train and test. Five different activities are recorded by Doppler radar. The number of clusters $K$ is estimated using four metrics. Since the raw data are in high-dimension, unsupervised feature extraction methods are employed to reduce the feature space dimension. The  feature extraction is followed by clustering and recognition using K-Means and K-Medoids. Comparison of the proposed two methods for feature extraction - local DCT-based method and local entropy-based method is performed with conventional methods based on CVAE, CAE, PCA, and 2DPCA. The existing CVAE method has not been deployed in previous Doppler radar studies.


\subsection{Dataset Description}
\begin{figure}[h]
\centering
\includegraphics[scale=0.43]{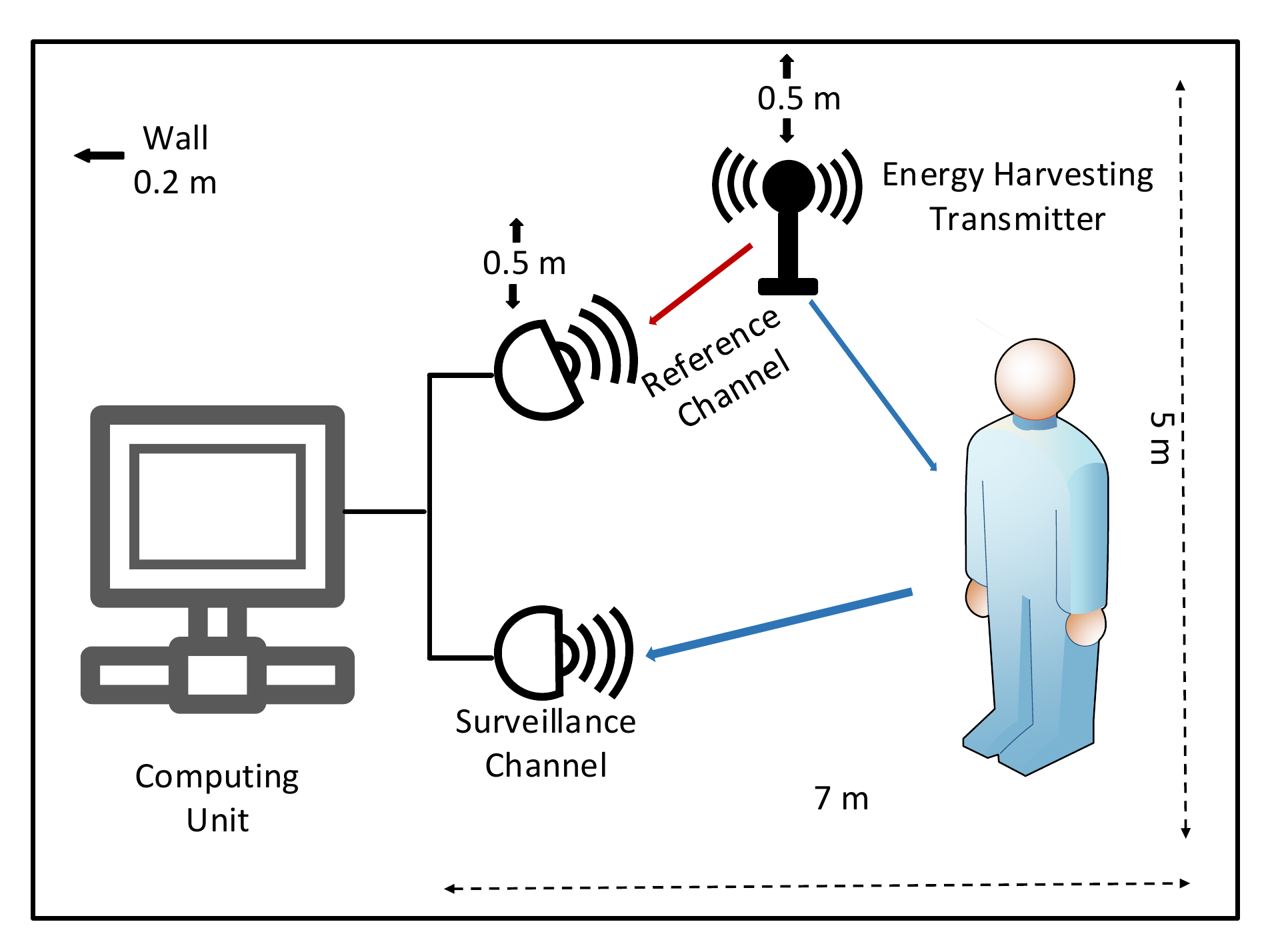}
\caption{Experimental layout.}
\label{fig:layout}
\end{figure}

The Doppler-spectogram dataset is collected in the University of Bristol laboratory. The laboratory experiment layout is shown in Fig. \ref{fig:layout} (a 7 m $\times$ 5 m room). The radio source used in this experiment is an Energy Harvesting transmitter (TX91501 POWERCASTER) working on 915 MHz ISM band with 30 dBm DSSS signal. The passive radar is a two-channel software defined radio (SDR), which is built on two synchronized NI USRP 2920s. Both channels are connected with directional antennas. The reference channel is 1 m apart from the transmitter, while the surveillance channel is pointed to the subject. The Cross-Ambiguity Function (CAF) which is the Fourier of cross-correlated reference and surveillance signals is used to 2D range-Doppler plot. From each range-Doppler plot, the range column which contains the detected subject is extracted to form up the Doppler spectrogram. More details can be found in Section 3 in \cite{Doppler:Healthcare} or Section III in \cite{Doppler:Pilot}.

Four participants (one female and three male) volunteered for capturing activities. This dataset consists of five activities: (1) walking, (2) running, (3) jumping, (4) turning, and (5) standing. Each activity is repeated 10 times by each subject. There exist 40 samples for each activity or 200 samples totally. 

\begin{figure*}\centering
\includegraphics[width=1\linewidth]{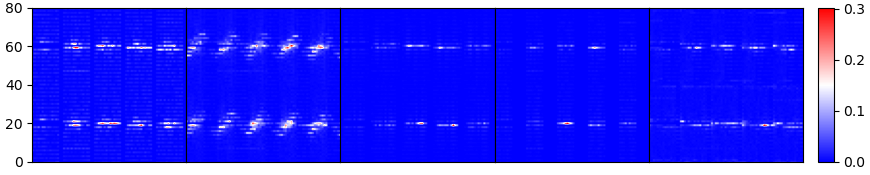}
\caption{Example of an $80 \times 80 = 6400$ image for each activity: (1) walking, (2) running, (3) jumping, (4) turning, and (5) standing.}
\label{fig:rawcombined}
\vspace{-1.5em}
\end{figure*}

A pre-processed Doppler radar dataset is used in this study. The total number of features per sample is 6400 = (2 directions $\times$ 100 Doppler bins $\times$ 32 time index). Furthermore, the Doppler radar data is normalized, which corresponds to the fact that all features are represented by real values in the range of (0, 1). Considering the 3-dimensionality of the Doppler radar data, it is then vectorized $2\times100\times32$, which results in $6400$. In order to transform it to 2D maps, $80\times80$ reshaping is applied. The reason for converting the vectorized Doppler radar data into 2D maps, is to apply image analysis strategies for quantification of local variation and patterns in the image. Fig. \ref{fig:rawcombined} is the micro Doppler signature for human activities.

The Python libraries used for data pre-processing are pandas (version 1.0.5) and numpy (version 1.19.1). In terms of machine learning for feature extraction and clustering, scikit-learn (version 0.22.2) and scikit-learn-extra (version 0.1.0b2) modules are applied. The CAE and CVAE are implemented and run with Keras (version 2.2.4) and Tensorflow (version 2.2.0). The visualisation results are implemented with matplotlib (version 3.2.2).

\subsection{Number of classes estimation}
Considering the unsupervised scenario in this study, the number of classes/clusters is unknown. In order to estimate the correct number, a number of techniques are applied including Elbow method, Silhouette analysis, Davies-Bouldin score and Dunn's index using K-Means clustering.

\subsubsection{Elbow method}
\begin{figure}[h]
\centering
\includegraphics[scale=0.44]{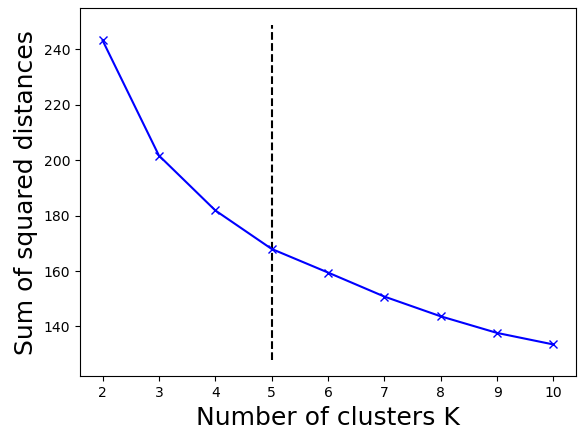}
\caption{An Elbow test used to determine the number of clusters $K$.}
\label{fig:elbowtest}
\end{figure}

The Elbow method is a heuristic technique for clusters number estimation \cite{Elbow:Source}, \cite{Elbow:Example}. The overall goal for the method is to maximize the inter-class variability and minimize the intra-class variability. In this study, the data samples are denoted as $D = D_{1}, D_{2},...,D{n}$. The number of clusters is $K$ and their centroids are given by $\omega_{1}, \omega_{2},...,\omega_{K}$. The distortion $J$ is used to measure the effectiveness of the method:

\begin{equation}
    J(K, \omega) = \frac{1}{n}\sum_{i=1}^{n}(\min_{j=1}^{K}(D_{i}-\omega_{j})^2)
\end{equation}
In this study, the candidate numbers of clusters $K=2, 3,...,10$ are selected for K-Means clustering, which is described later in this section. The Elbow method computes the sum of squared errors for the data samples in each cluster. As the number of clusters increases, $J$ becomes smaller. However, the best value of $J$ is the point, where a further increase to the number of clusters does not change the within-cluster sum of squares significantly. However, a further increase would result in over-clustering. The decrease trend of $J$ is noticeable before reaching the actual number of clusters $K$ and becomes smoother afterwards. The Elbow method is a visualisation tool, and the graph shows a noticeable decline when the curve approaches the actual $K$. Therefore, the decline becomes smoother after exceeding $K$. Fig. \ref{fig:elbowtest} illustrates the Elbow test for this data.

As it can be observed, the selected number of clusters $K = 5$, which is the actual number of clusters for this study. However, detection of this bend point is ambiguous in some cases. Therefore, additional techniques are considered in this study. 

\subsubsection{Silhouette analysis}
Silhouette analysis is one of the most commonly used techniques for number of clusters estimation\cite{Silhouette:Source}. The method is given as:

\begin{equation}
    Silhouette = \frac{1}{n}\sum\limits_{i=1}^{n}\frac{b(D_{i})-a(D_{i})}{\max\{a(D_{i}),b(D_{i})\}}
\end{equation}
where $a(D_{i})$ is the average distance between data point $D_{i}$ and the remaining data points in its own cluster. The minimum average distance between data point $D_{i}$ and all other clusters is denoted with $b(D_{i})$. The Silhouette coefficient aims to show the suitability for data point $D_{i}$ to belong to a particular cluster. The score is within the range of (-1, 1), where a lower value refers to overlapping clusters. On the other hand, a higher value suggests well-separated clusters.





\subsubsection{Davies-Bouldin index}
The Davies-Bouldin index is a clusters estimation method concerned with identifying clusters, which are distinct from each other \cite{Davies:Source}. The measure is given by:

\begin{equation}
    Davies = \frac{1}{K}\sum\limits_{i=1}^{K}\max\limits_{i\neq{j}}\frac{\Delta(C_{i})+\Delta(C_{j})}{\lambda{(\omega_{i},\omega_{j})}}
\end{equation}
where $\Delta(C_{i}) = \sum\limits_{D_{i}\in C_{i}}^{n} ||D_{i}-\omega_{i}||^{2} $, which is the distance between each Doppler sequence and the centroid of the corresponding cluster. The distance between cluster centroids is given by $\lambda{(\omega_{i},\omega_{j})} = ||\omega_{i}-\omega_{j}||^{2}$. Then, the term for maximization is computing the ratio of within-cluster to between-cluster distances for the $i^{th}$ and the $j^{th}$ clusters. This is computed for all combinations of cluster $i$ and other clusters. Then, the maximum value is found for each $i$. Among all combinations of clusters, the closest clusters with largest spreads have the maximum ratio. That is the worst scenario. The desire is to minimize the overall average of the worst scenario ratios. Therefore, unlike previous methods, a smaller value for Davies index is desirable.

\subsubsection{Dunn's index}
Dunn's index is one of the most popular and oldest techniques in the literature \cite{Dunn:Source} for number of clusters estimation. The overall aim is to minimize the intra-cluster distance and maximize the inter-cluster distance. It is given by:

\begin{equation}\label{eq:Dun}
    Dunn = \frac{\min\limits_{1\leq i <j\leq K} \lambda(\omega_{i},\omega_{j})}{\max\limits_{1\leq k \leq K}\Delta(C_{k})}
\end{equation}
where $\lambda(\omega_{i},\omega_{j})$ is the distance between clusters $C_{i}$ and $C_{j}$. The intra-cluster distance of a single cluster is given by $\Delta(C_{k})$. The minimization in the numerator finds the Euclidean distance of the two closest clusters. On the other hand, the maximisation in the denominator, finds the Euclidean distance of the samples to the centroid of the cluster with the highest dispersion. Therefore, for optimum cluster number, the numerator will be the maximum value among all other candidate number of clusters, while the denominator will be the lowest, resulting a peak over the heuristic search.
\bigskip

\begin{figure}[h]
\centering
\includegraphics[scale=0.52]{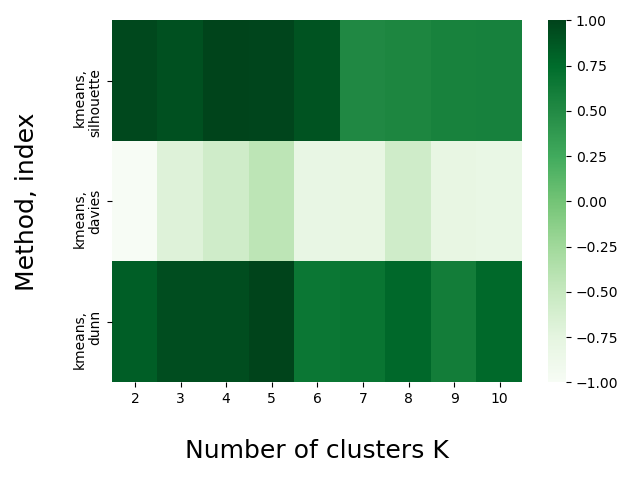}
\caption{Silhouette score, Davies score and Dunn's index used to determine the number of clusters $K$.}
\label{fig:validclust}
\end{figure}

Similarly to the Elbow method implementation, a candidate set of clusters is given $K=2,3,...,10$ for K-Means. The heatmap in Fig. \ref{fig:validclust} shows the results for each heuristic technique. It is important to note that the value for the inverse of Davies index is used. Hence, the highest values corresponding to the darkest colors show the best fit for the clusters. The selected number of clusters is $K=5$ for Davies and Dunn's index, while the Silhouette coefficient is very similar for $K=4$ and $K=5$. Overall, $K=5$ is selected as the most optimum number of clusters by the majority of methods.

\subsection{Feature extraction methods}
Traditionally, researchers rely on the empirical knowledge for the human activity recognition using micro-Doppler signatures. The typical examples include average torso velocity, period or duration of the activity cycle, upper and lower envelope variances in \cite{SVM:Doppler,Traditional_feature_2}. The empirical knowledge presents intuitive relation between the Doppler signature and feature, however, it is not suitable for the datasets with uncontrolled or unknown conditions. Later on, the data-driven micro-Doppler recognition approaches are proposed and proven excellent performance in \cite{Deep_feature_1,Deep_feature_2,Deep_feature_3}. These approaches treat the micro-Doppler plots as time-spectrogram and range-Doppler time points cluster respectively. In this work, we will explore two new local DCT-based and local entropy-based methods as well as convolutional filter-based and variation-based projection methods for feature extraction. The two new methods and convolutional filter-based strategies are superior in terms of accuracy. While the convolutional strategy is superior in accuracy, the training time of the new methods is considerably less than the convolutional strategy.

\subsubsection{The proposed local DCT-based method}
\begin{figure*}[h]
\centering
\includegraphics[width=\linewidth]{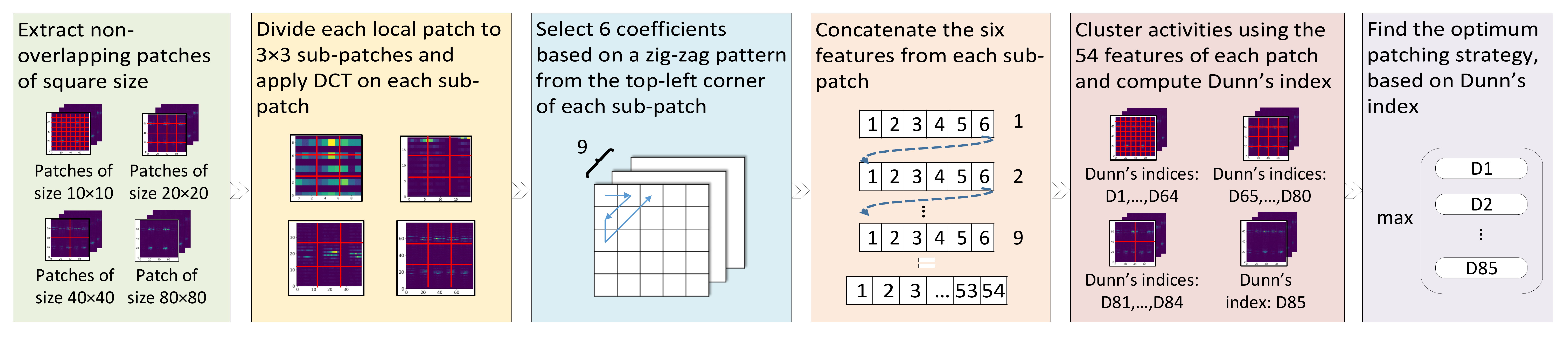}
\caption{Flowchart of the proposed local DCT-based method.}
\label{fig:dct_flowchart}
\end{figure*}
The first proposed method for feature extraction is based on applying 2D DCT on local areas of the 2D map of Doppler radar data. In general, 2D DCT is used to transform 2D images from the spatial domain to the frequency domain \cite{DCT}. The sharp changes or smooth variations in the images correspond to high frequencies or low frequencies in DCT domain respectively. In fact, the frequency information of the images is sorted by DCT transform. The DCT coefficient in the top-left corner of the output 2D DCT matrix corresponds to the lowest frequency of zero and the frequencies increase toward the bottom-right corner. Depending on the type of images, the higher energies appear in different coefficients. The most valuable information is usually in small fractions of the DCT images. Therefore, DCT allows selection of a limited number of features that reduces the dimension of the feature space. The method has been used previously for extracting features from micro-Doppler radar for human activity recognition \cite{DCT:Doppler}. The mathematical notation of DCT is given as:
\begin{equation} \label{equ:2D_DCT}
    \resizebox{.91\hsize}{!}{$F[p,q] = \frac{2}{\sqrt{uv}}a_{p}a_{q} \sum\limits_{i=0}^{u-1}\sum\limits_{j=0}^{v-1}  f[i,j]\cos\frac{\pi(2i+1)p}{2u}\cos\frac{\pi(2j+1)q}{2v}$}
\end{equation}
where, $F[p,q]$ values are the DCT coefficients at row $p$ and column $q$. In addition, $f[i,j]$ is the element in row $i$ and column $j$ of the image matrix, where $u=80$ and $v=80$, and $0 \leq p \leq u-1$ and $0 \leq q \leq v-1$. Moreover, $a_{p} = \frac{1}{\sqrt{2}}$ if $p=0$ and it is 1 otherwise. Similarly, $a_{q} = \frac{1}{\sqrt{2}}$ if $q=0$ and its value is 1 otherwise.

Considering the local variations of the original $80\times80$ images, in this paper, a systematic search algorithm is proposed in order to find the best strategy for applying 2D DCT. This includes applying DCT on different local areas of the $80\times80$ 2D maps, using various patch sizes. The aim is to identify the optimum patch size giving the best clustering results. As illustrated in Fig. \ref{fig:dct_flowchart}, first the 2D images are divided into various square shape local patches of different sizes. The square-sized local patches are non-overlapping. Four sizes of local patches are considered - $10\times10$, $20\times20$, $40\times40$ and the original $80\times80$ 2D map. Second, each patch is divided into $3\times3$ sub-patches. Third, 2D DCT is applied to each local sub-patch and the resulting 2D map of the DC coefficient's amplitude is used for feature selection. Third, six coefficients are extracted according to a zig-zag pattern from the top-left corner of each sub-patch, allowing to extract features from all local areas of the Doppler profiles. Finally, the six features from each of the nine sub-patches are concatenated to form a feature vector of size $9\times6=54$. This generates $54$ features for each local patch. Then, the activities are clustered using the 54 features of each patch and the Dunn's index is computed. The highest Dunn's index discovers the most optimum local patch for the 2D DCT analysis.


The reason for using the Dunn's index in this algorithm is that it describes the quality of the resulting clusters. It quantifies an easily interpretable metric based on the worst clusters of a clustering scenario. As shown in (\ref{eq:Dun}), in its numerator there is the minimum between-cluster distance and the denominator is the maximum within-cluster distance. Then, a high Dunn's index shows a good clustering quality. The use of Dunn's criterion rather than clustering accuracy allows parameter selection for the unsupervised framework.

\subsubsection{The proposed local entropy-based method}
Considering the $80\times80$ images, a texture analysis method based on entropy is proposed to quantify the patterns of different activities profiles. Entropy is a statistical measure of randomness and is formulated based on Shannon's equation \cite{Entropy:Shannon} as follows:
\begin{equation}
H(\rho)=-\sum_{i=1}^{b}\rho_{i}\log(\rho_{i})
\end{equation}
where $\rho_{i}$ is the normalized histogram counts. It is calculated based on the histogram of the image. $b$ is the total number of histogram bins.

\begin{figure}[h]
  \centering
  \begin{subfigure}{.3\linewidth}
    \centering
    \includegraphics[width = \linewidth]{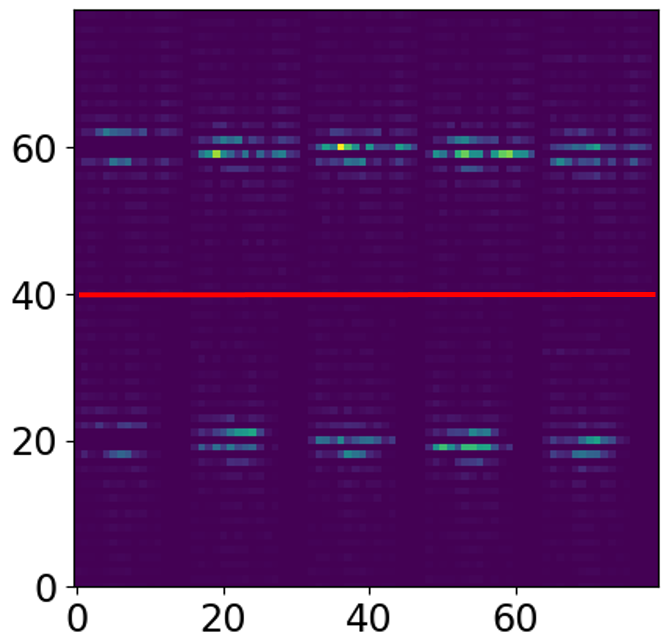}
    \caption{First strategy}
  \end{subfigure}%
  \begin{subfigure}{.3\linewidth}
    \centering
    \includegraphics[width = \linewidth]{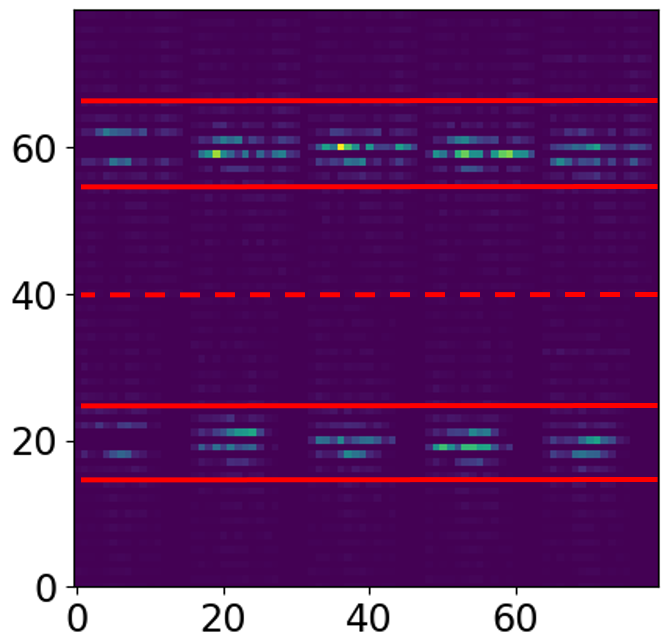}
    \caption{Second strategy}
  \end{subfigure}%
  \begin{subfigure}{.3\linewidth}
    \centering
    \includegraphics[width = \linewidth]{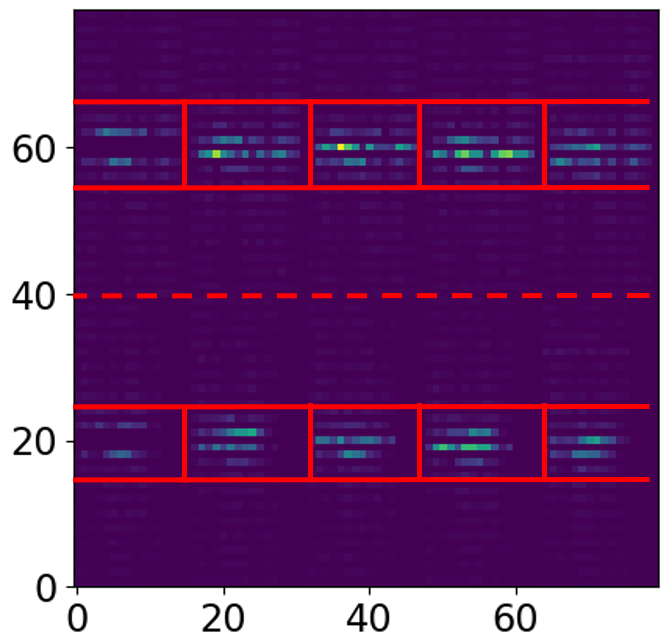}
    \caption{Third strategy}
  \end{subfigure}
  \caption{The three patching strategies for entropy analysis.}
  \label{fig:entropypatch}
\end{figure}

\begin{figure*}[h]
\centering
\includegraphics[width=\linewidth]{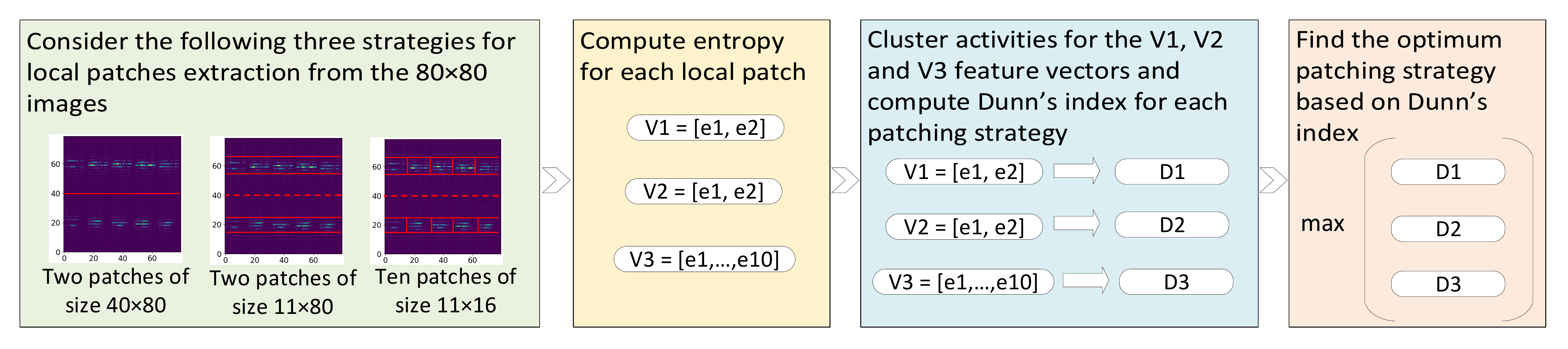}
\caption{Flowchart of the proposed local entropy-based method.}
\label{fig:entropy_flowchart}
\end{figure*}

Depending on the variations of colors in local image area, the entropy can change. If most pixels in an image are similar with a low level of variations, the entropy will be small. On the other hand, if the level of color variation is high in an image, the entropy increases. Therefore, depending on the location of the analysis window, the entropy value can change. Since the patterns and color intensities vary for different activities, a careful selection of local patches can generate different entropy values suitable for discrimination of the activities.
Based on the observed changes in Fig. \ref{fig:rawcombined}, three patching strategies are considered so that, the selected image areas for entropy analysis are narrowed down systematically. The three local patchings strategies are illustrated in Fig. \ref{fig:entropypatch}. Then, similar to the local DCT-based analysis, the Dunn's index is used to evaluate the quality of the clustering results based on the entropy features. That allows identifying the best patching strategy. The steps of the proposed method are outlined in Fig. \ref{fig:entropy_flowchart}.

\subsubsection{Convolution filter-based methods}
In this section, a description of CAE is provided by considering its drawbacks. Then, CVAE is introduced, which overcomes the drawbacks of the previous architecture.
\paragraph{Convolutional Autoencoder (CAE)}
Autoencoders (AEs) are unsupervised neural networks, which can be used for feature extraction. Their architecture consists of two components: an encoder and a decoder \cite{CAE:Source}. AEs are commonly used for data denoising \cite{CAE:Denoising}, anomaly detection \cite{CAE:Anomaly} and image generation \cite{CAE:Images}. The encoder learns the latent attributes of the input data $x$ and transforms it to a lower dimensionality representation $z$. On the other hand, the decoder aims to reconstruct $x$ given $z$. The implementation of CAE only contains a reconstruction loss, which needs to be minimized.

In regards to disadvantages of the discussed architecture, the CAE learns local parameters for each data point. This avoids any statistical strength to be shared across all data points. Hence, this may result in overfitting due to the inability of the model to generalise. In addition, the CAE architecture includes only a reconstruction loss and it lacks any regularisation term as seen in CVAEs. This leads to data points of the same group/class to be given different representations, which are often meaningless.

\begin{figure*}[h]
\centering
\includegraphics[width=0.95\linewidth]{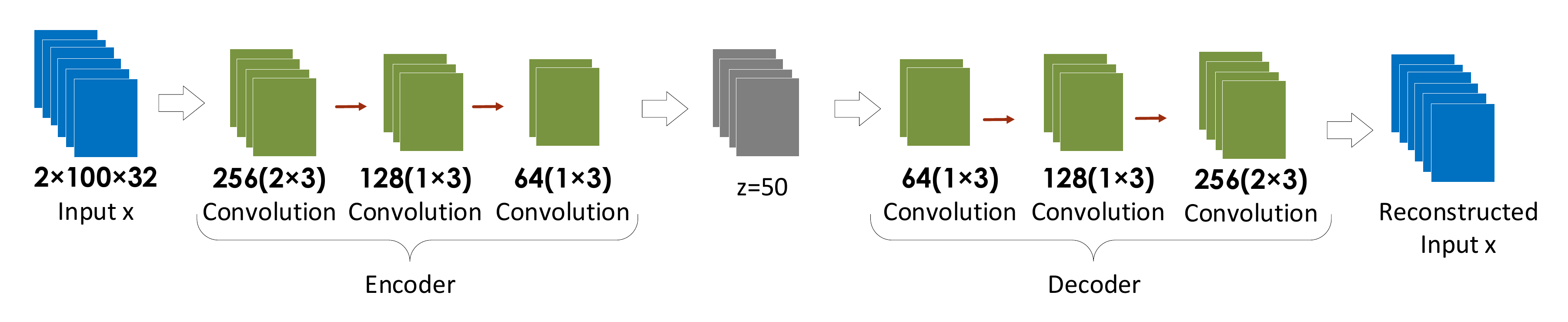}
\caption{The CAE architecture.}
\label{fig:cae}
\end{figure*}

In this work, a deep CAE is used with three hidden layers for the encoder and decoder as illustrated in Fig. \ref{fig:cae}. As it can be seen, the shape of the input data $x$ is $2\times100\times32$ and the retained number of latent variables $z$ is 50. The structure of the encoder is symmetric to the decoder's structure. In regards to the hidden layers, the first convolutional layer in the encoder and the third convolutional layer in the decoder have 256 filters with size $2\times3$. The stride for these two layers is (1, 2) referring to height and width. The second convolutional layer in the encoder and the decoder have 128 filters with size $1\times3$ and the stride is of shape (1, 2). Considering the third layer in the encoder and the first layer in the decoder, they have 64 filters with size $1\times3$. Their stride is of shape (2, 1). The robustness of CAE is validated in Section \ref{SecIV:Results}. The convolutional layers used for this architecture incorporate a ReLU activation function. The decoder's task is to reconstruct $x$ given $z$, which is evaluated with the reconstruction loss. Based on this architecture, the encoded features $z$ are used for clustering the activities.

\paragraph{Convolutional Variational Autoencoder (CVAE)}
\begin{figure*}\centering
\includegraphics[width=0.95\linewidth]{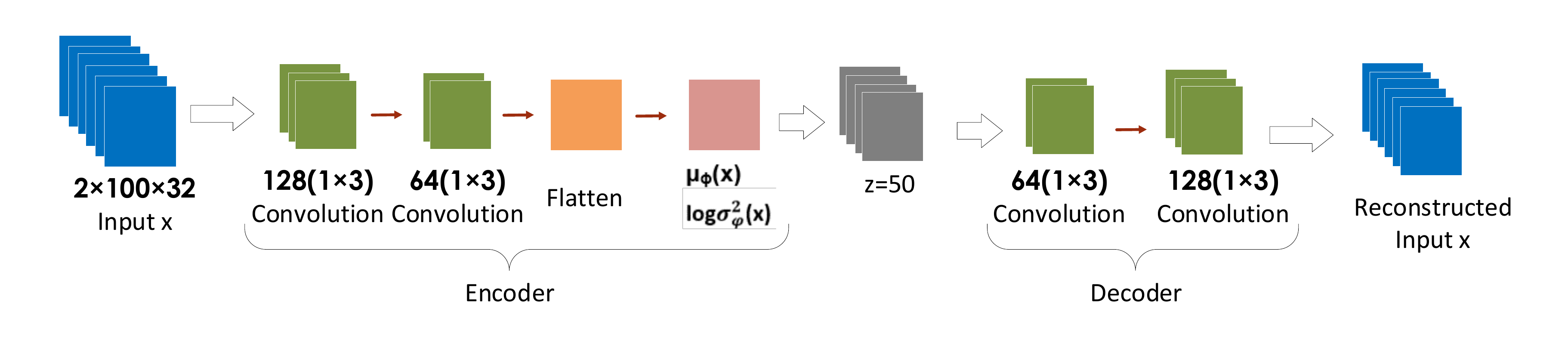}
\caption{The CVAE architecture.}
\label{fig:vae}
\vspace{-1.5em}
\end{figure*}
CVAEs are generative models defined in \cite{VAE:Proposed}, which are commonly used for dimensionality reduction \cite{VAE:FE}, data augmentation \cite{VAE:DA}, and reinforcement learning \cite{VAE:RL}. Considering Doppler radar data, CVAEs have been used for synthetic data generation \cite{CVAE:Augmentation}. CVAEs contain two main modules: an encoder, referred to as recognition model or inference model, and a decoder, also defined as generative model \cite{VAEs:Introduction}. The purpose of the encoder is to learn the stochastic mappings of the observed input space $x$ with a rather complicated distribution and transform it from its original high-dimensional space (6400 dimensions in this case) into a much lower latent representation $z$ with a relatively simple distribution. Then, the output of the recognition model $z$ is the input of the decoder. The decoder aims to reconstruct the original input $x$ from the reduced latent information. In a traditional autoencoder, the latent representation consists of single-valued outputs for each feature. CVAEs assume that the dimensions of $z$ cannot be interpreted with simple variables. Instead, CVAEs introduce a probability distribution for the samples of $z$, which is commonly a Gaussian distribution \cite{VAEs:Tutorial}. The use of a single reconstruction error in encoder-decoders might result in encoding some meaningless content. That results in overfitting and therefore the latent space should be regularised. Contrarily to CAE, in the CVAE architecture, the loss function includes a reconstruction term and a "regularisation" term. The latter term is developed by enforcing the probabilistic distribution of the encoded space to be close to a Standard Normal distribution. This is expressed as the Kullback-Leibler (KL) divergence. The KL divergence quantifies the divergence of the latent space distribution, denoted as $q_{\theta}(z|x)$ in (\ref{eq:VAE}) and the standard normal distribution $p(z)$:
\begin{equation}\label{eq:VAE}
    l_{i} = -E_{z\sim q_{\theta}(z|x_{i})}[log p_{\phi}(x_{i} | z)] + KL(q_{\theta}(z|x_{i}) || p(z))
\end{equation}
where $p_{\phi(x_{i}|z)}$ describes the generative probability of the reconstructed output given the encoded variable $z$. The distribution of the encoded variable $z$ given the input $x_{i}$ is denoted as $q_{\theta}(z|x_{i})$. In addition, $\theta$ and $\phi$ are parameters of the distribution.

The architecture of the developed CVAE model is illustrated in Fig. \ref{fig:vae}. Similar to the CAE architecture, the input for CVAE is of shape ($2\times100\times32$) corresponding to height, width, and depth. The ReLU activation function is also used similarly. The structure of the encoder is again symmetric to the decoder's structure. The first convolutional layer in the encoder and the second convolutional layer in the decoder have 128 filters with size $1\times3$. Their stride is of shape (1, 2) corresponding to height and width. The second layer in the encoder and the first layer in the decoder have 64 filters with size $1\times3$. Their stride is of shape (2, 2). The encoding part does not forward the direct latent values to the decoder. Instead, mean $\mu_{\phi}$ and variance $\log\sigma_{\phi}^{2}x$ vectors of the latent features are its output. These parameters are enforced to be close to a standard normal distribution, which is measured by the regularisation term. Finally, both $\mu_{\phi}$ and $\log\sigma_{\phi}^{2}x$ are sampled to produce the compressed latent space representation $z$ with the specified number of features. Considering the latent representation $z$, the decoder aims to reconstruct $x$. The effectiveness of this operation is evaluated with the reconstruction loss.

\subsubsection{Variation-based projection techniques}
\paragraph{Principal Components Analysis (PCA)}
1D-PCA is known to be one of the most common linear techniques for unsupervised feature extraction \cite{PCA:Tutorial}. PCA finds the directions of main variations of data in the original high dimensional space, and projects data along those directions into a smaller sub-space. Based on this linear projection, the dimensionality of data is reduced. In this paper, the Doppler radar data with 6400 variables are used for PCA analysis. Based on a weighted linear combination of these features, the main directions of variations of data are calculated. In this paper, the $s$ number of the first few eigen vectors, explaining 95\% of data variations, is used to transform the original high dimensional data $D_{n\times m}W_{m\times s}=Z_{n\times s}$. The first principal component (PC1) usually retains the highest variance, which allows a smaller number of PCs to be selected.


\paragraph{2D Principal Components Analysis (2DPCA)}
PCA requires the 2D image matrix to be transformed to a 1D image vector. This often leads to a high-dimensional image vector. Therefore, the size of the covariance matrix $V$ is extremely large. Logically, it becomes difficult to evaluate the covariance matrix considering the small number of training examples. 2-dimensional PCA proposed in \cite{2DPCA:Source} allows the covariance matrix to be calculated on the 2D images of size $80\times80$. Hence, this corresponds to its smaller size, which has two main advantages. Less computation time is required and the covariance matrix is more accurately evaluated. The covariance matrix is given by:
\begin{equation}
\displaystyle \mathrm{V}=\frac{1}{n}\sum_{i=1}^{n}(\mathrm{X}_{i}-\overline{\mathrm{X}})^{T}(\mathrm{X}_{i}-\overline{\mathrm{X}}) \
\end{equation}
where $n$ is the number of training samples and $\overline{\mathrm{X}}=\frac{1}{n}\sum_{i=1}^{n}X_{i}$ is the average training image. It also has the size $80\times80$.

More specifically, 2DPCA computes the covariance matrix only for the row or column dimension only. That is because the initiall data is not vectorized to include all features. Considering the $80\times80$ 2D images in this study, only the columns are used for computing the covariance matrix. This process is followed by eigen decomposition of the covariance matrix. PCs retaining most of the variance are then selected. Similar to PCA, the first PC retains most of the variance.



\subsection{Clustering methods}
Two clustering methods are used for grouping the data samples. Both of the methods are distance-based: K-Means and K-Medoids.
\subsubsection{K-Means}
K-Means clustering is one of the most commonly used techniques for unsupervised learning \cite{K-Means:Introduction}. The K-Means method considers the number of groups or clusters $K$ is known and it aims to group the data points based on their distances. In this work, the Euclidean distance is used. The overall goal for the clustering method is to group the data points $D_{1}, D_{2},...,D_{n}$ in clusters by minimizing the intra-cluster distances. Intuitively, the distances between data points from different clusters should be maximized. During training, K-Means outputs the cluster centres $\omega_{k}$, where $k = 1, 2,..., K$, or also known as centroids. The assignment of a new data point to one of the clusters is such that, the sum of the squared distances between the data point and all cluster centroids $\omega_{1}, \omega_{2},...,\omega_{k}$ are computed. Then, the sample is assigned to the cluster, where the corresponding distance is minimum. The objective function for K-Means, which specifies the sum of squared distances of each data point $D_{i}$ to cluster $k$, is defined as follows:

\begin{equation}
    J = \sum_{i=1}^{n}\sum_{k=1}^{K}r_{ik}||D_{i} - \omega_{k}||^2
\end{equation}
where $r_{ik}$ is a binary function indicating the assignment of data point $D_{i}$ to cluster $k$. If $D_{i}$ is assigned to cluster $k$, the binary indicator $r_{ik} = 1$, and 0 otherwise:

\begin{equation}
    r_{ik}= 
\begin{cases}
    1,& \text{if } k = \arg\min_{j}||D_{i} - \omega_{j}||^2\\
    0, & \text{otherwise}
\end{cases}
\end{equation}
The overall aim is to minimize the $J$ for values $r_{ik}$ and $\omega_{k}$. The procedure can be achieved by iterative optimization with respect to $r_{ik}$ and $\omega_{k}$. In the first phase, the $\omega_{k}$ is fixed, while the goal is to optimize $r_{ik}$. The same notion is applied to the second step as $r_{ik}$ is fixed and the focus is on the optimization of $\omega_{k}$. The entire process corresponding to Expectation-Maximization algorithm is repeated until convergence.

\subsubsection{K-Medoids}
K-Medoids is a clustering method based on distances analysis, which has shown better performance for noisy and problematic data than K-Means \cite{K-Medoids:ESL}. Similar to K-Means, this method considers the number of groupings or clusters $K$ is known initially and $K<n$, where $n$ is the number of data points. In contrast, K-Medoids considers a data sample for the centroid or the medoid, which is not the case for K-Means. 
In the first step of K-Medoids, the algorithm aims to find a data point $D_i$ in a cluster $C(i)=k$ that is in minimum distance to the remaining observations in the cluster $D_i'$. This distance is denoted as $||D_{i}-D_{i'}||^{2}$ and the minimisation is shown in (\ref{equations:medoids}):

\begin{equation}\label{equations:medoids}
    i_{k}^{*} = \argmin\limits_{i:C(i)=k}\sum\limits_{C_{i'}=k}^{n} ||D_{i}-D_{i'}||^{2}
\end{equation}
Then, the output index $i_{k}^{*}$ is used to find a new centroid or medoid, defined as $\omega_{k} = D_{i_{k}^{*}}$, $k=1,2,...,K$ for all clusters. 

The second step of the method is to minimise the total error by re-assigning each data sample to the closest centroid. The clusters centroids are given ${\omega_{1},\omega_{2},...,\omega_{K}}$.

\begin{equation}
    C(i) = \argmin\limits_{1\leq k \leq K} ||D_{i}-\omega_{k}||^{2}
\end{equation}
Finally, step 1 and step 2 are iterated until the algorithm converges to the optimum centroids.

\subsection{Visualisation techniques for high-dimensional data}
In this paragraph, three widely-known techniques for manifold learning are defined. The focus is on transforming the very high-dimensional space in this dataset ($D_{n \times m} = D_{n\times6400}$) to a 2-dimensional space. Manifold learning methods are known to map closely correlated data samples in similar positions, while the gap in the low dimensional space increases if the samples are non-similar. Hence, comparison measures will be extracted from this analysis, which can be useful for projects concerned about high-dimensional data visualisation.

\subsubsection{t-Distributed Stochastic Neighbour Embedding (t-SNE)}
T-SNE is a non-linear dimensionality reduction method, which has gained attention for its superior ability to visualise high-dimensional data by transforming it to a two or three-dimensional space \cite{t-SNE:Introduction}. This method assigns each data point in a low-dimensional location by aiming to preserve the significance of the original information. Unlike linear techniques such as PCA and SVD, t-SNE aims to keep similar data points in close locations in the low-dimensional space. The superiority of t-SNE in comparison with other dimensionality reduction methods is the ability to preserve the local structure of the data as well as global information such as clusters. T-SNE has been recently compared with PCA for visualisation where the former achieved better visualisation \cite{t-SNE:Comparison}. The steps for the t-SNE transformation are described below:

\begin{enumerate}
    \item The Doppler sequences $D_{1}, D_{2},...,D_{n}$ are initially in their original 6400-dimensional space. T-SNE begins with determining the similarity between the data samples. This is performed by computing their distances. Euclidean distances are used in this study.
    \item The Euclidean distances are converted to probabilities describing normal distributions so that, similar data samples have close values. On the other hand, dissimilar points have distinct similarity values. The similarity scores are calculated for each data points pair $D_{ij}$, where a similarity matrix is obtained based on probabilities $p_{ij}$.
    \item The data samples are projected in a random order to the low dimensional space first. This results in a mismatch with cluster patterns of data in the original domain initially. The aim for t-SNE is to re-position the data samples in the new low dimensional space, such that the same clustering patterns of the high-dimensional space to be preserved.
    \item Then, the Euclidean distances between the data samples are calculated in the lower dimensional space. Similarly, the distances are converted to a t-Distribution (e.g. $q_{ij}$ for the two data points $D_i$ and $D_j$). t-Distribution is similar to normal distribution, but with taller tails. The taller tails of t-Distribution prevent dissimilar data points to be positioned in close locations of the lower dimensional space. The samples in lower dimension are re-positioned using these probabilities resembling distances. The re-positioning is performed based on the the two probabilities $q$ of low dimension and $p$ of high dimension.
    \item t-SNE uses KL divergence to optimise the similarity of the distributions described by $q$ to those described by $p$. This can be interpreted as a constant comparison of the samples distances in the lower-dimension  to their distances in the original high-dimension. Then, re-positioning will be improved iteratively, as the similarity matrix of probabilities $q$  is optimized using the original similarity matrix $p$.
\end{enumerate}

\subsubsection{MultiDimensional Scaling (MDS)}
MDS is a non-linear dimensionality reduction technique. It can be used for visualization of high dimensional data in low dimensional space. It preserves the actual distances of original samples in the low dimensional space. MDS considers dissimilarities of sample pairs contrary to other methods, which are concerned with similarities. Given the set of observation $D_1,D_2,...,D_n\in\mathbb{R}^m$, $d_{ij}$ is the dissimilarities e.g. the Euclidean distance of two samples $D_{i}$ and $D_{j}$ so that, $d_{ij}=||D_{i}-D_{j}||^{2}$. MDS seeks $z_1,z_2,...,z_n\in\mathbb{R}^k$, so that $k<m$. A so called stress function is minimized for this aim \cite{MDS:1},\cite{MDS:2},\cite{MDS:ESL}:

\begin{equation}\begin{medsize}
    Stress_{(z_{1},...,z_{n})} =\sum\limits_{i\neq{j}}^{n}(d_{ij}-||z_{i}-z_{j}||)^{2}
\end{medsize}\end{equation}
where $||z_{i}-z_{j}||$ is the Euclidean distance between $z_{i}$ and $z_{j}$. Then the pairwise distances are preserved in the lower dimensional representation. A gradient descent algorithm is used to minimize the stress function and find the components in the low dimension \cite{MDS:ESL}. 
MDS transformation is monotone increasing with the increasing dissimilarities. The same notion is applied to growing similarities data, which decreases the transformation. Hence, similar object pairs are positioned closely in the transformed space, while objects with dissimilarity are distinguished with larger distances.

\subsubsection{Locally Linear Embedding (LLE)}
LLE is a non-linear dimensionality reduction method proposed in \cite{LLE:Introduction}. The method is concerned with preserving the global structure of the data based on an underlying manifold. The data are represented by $n$ real-valued vectors $\Vec{X_{i}}$ in a high dimension $m$. $i$ is the index of a sample. Each data point $\Vec{X}_{i}$, is a member of a neighbourhood. Each neighbourhood consists of similar data points. Similar data points are expected to lie on a close locally linear patch of the smooth manifold. The $p$ nearest neighbours for each data point are defined by measuring the Euclidean distances. The local geometry of the patches can be characterised by linear coefficients. These linear coefficients are used to reconstruct each data point from its neighbours. The reconstruction loss is defined by:

\begin{equation}
    \epsilon(W) = \sum_{i=1}^{n}||\Vec{X_{i}} - \sum_{j=1}^{p}W_{ij}\Vec{X_{j}}||^{2}
\end{equation}
where $W_{ij}$ are the weights defined for data points reconstruction using the corresponding neighbours. The number of samples is given as $n$, while the number of neighbours is $p$. The computed weights $W_{ij}$ correspond to the contribution of a data point $\Vec{X}_{j}$ for reconstructing $\Vec{X}_{i}$. In order to ensure that $\Vec{X}_{i}$ is reconstructed only by its neighbours, the weight function $W_{ij} = 0$, if a sample $\Vec{X}_{j}$ does not belong to the same class. Another constraint to the reconstruction loss is that the sum of the weight matrix's rows should be one, $\sum\limits_{j=1}^{p}W_{ij} = 1$. This sum-to-one constraint makes the weights invariant to translation of the data points and their neighbors. The weights are also invariant to rotation and scaling. The minimisation of the loss function, allows computation of the weights $W$. They characterize the intrinsic geometric properties of each neighborhood. 

Using the weights, $W_{ij}$, it is possible to project each high-dimensional data point $\Vec{X_{i}}$ to vector $\Vec{Y}_{i}$ of the lower representation based on another reconstruction cost function. Having the $W_{ij}$ fix, the aim is to minimise the embedded cost function to optimise the low d-dimensional coordinates ($d<m$) :


\begin{equation}
    \Phi(Y) = \sum_{i=1}^{n}||\Vec{Y}_{i} - \sum_{j=1}^{n}W_{ij}\Vec{Y}_{j}||^{2}
\end{equation}
\section{Evaluation and results} \label{SecIV:Results}
In this section, the results obtained using the four groups of unsupervised feature extraction techniques are presented. These include the two proposed methods, namely, local DCT-based method and local entropy-based method. In addition, the existing convolutional filter-based, and variational-based projection methods are used for comparison.

For local DCT-based method, local entropy-based method, and 2DPCA, the inputs are $n$ reshaped images of size $80\times80$. In the case of CAE and CVAE, the inputs are $n$ number of 3D cubes of size $2\times100\times32$. While for PCA, the input data is $D_n\times6400$. In the case of PCA and 2DPCA, the selected number of eigen vectors preserves 95\% of the data variance.


Leave-one-subject-out cross validation (LOOCV) is used to avoid over-fitting. As listed in Section \ref{SecIII:Methodology}, four participants are included in the data. The four participants correspond to the four folds. The models are trained on three subjects data. Then, they are validated on unseen data from the remaining subject, which are not used for building the models. This is repeated for all four subjects. Hence, $Z_{tr}=150\times m$ is the training matrix and $Z_{ts}=50\times m$ is the matrix for testing, where the number of features $m$ varies for different models.

In addition to the activities clustering results, the three manifold learning methods t-SNE, MDS and LLE are compared in two scenarios. In the first scenario, they are used to transform the raw data features to a 2-dimensional space. Since CVAE encoded features obtained the most accurate clustering results, in the second visualization scenario, they are used for projection into a 2-dimensional space.

True labels are only used for model evaluation and illustration purposes. The order of the predicted labels by clustering is not necessarily consistent with the actual labels order. Therefore, the clustering accuracy is estimated by finding the best-matching pairs of clusters labels and true labels. Based on this, the predicted labels by clustering  are matched to their corresponding actual true labels. As such, a regular accuracy score function is used for calculating the accuracy.

\subsection{Local DCT-based analysis results}

\begin{figure}[h]
\begin{subfigure}{0.235\textwidth}
\includegraphics[width=\textwidth]{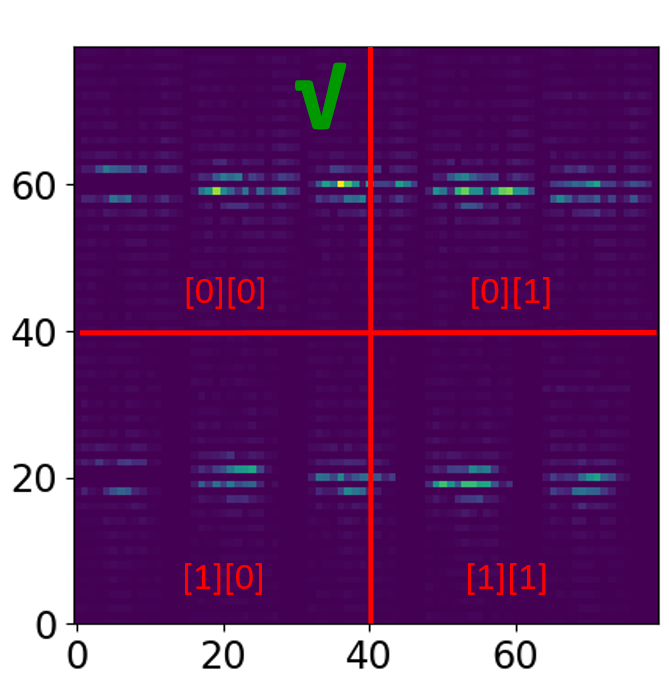}
\caption{}
\label{fig:dunnsize}
\end{subfigure}
\begin{subfigure}{0.235\textwidth}
\includegraphics[width=\textwidth]{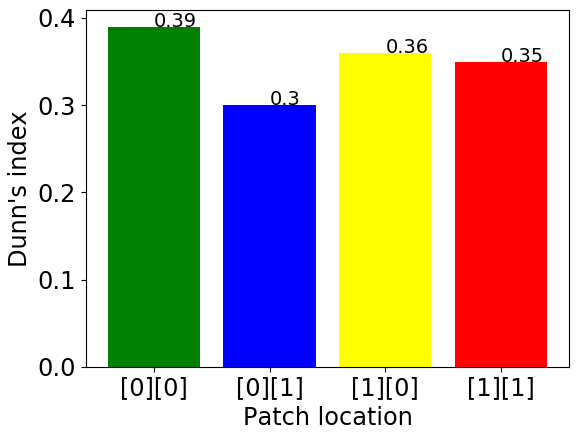}
\caption{}
\label{fig:dunnloc}
\end{subfigure}
\caption{(a) Illustration of the $40\times40$ local patches in the Doppler spectogram. The optimum patch for feature extraction is ticked in green color. (b) The corresponding Dunn's index for the four $40\times40$ local patches.}
\label{fig:dunneval}
\end{figure}

The proposed local DCT-based method extracts non-overlapping square-sized patches from the original $80\times80$ 2D map for analysis. Dunn's index is used for validating the method, which showed the highest values for the $40\times40$ local patches. In Fig. \ref{fig:dunnsize}, the four possible non-overlapping patch locations for this size are illustrated. Then, each $40\times40$ patch is divided into $3\times3$ sub-patches as was shown in Fig. \ref{fig:dct_flowchart} previously. Next, six DCT coefficients of the top-left zig-zag pattern are  selected from each sub-patch yielding a total of $9\times6=54$ DCT coefficients for each $40\times40$ patch. The features are then used for clustering. The resulting Dunn's indices are visualized and compared in Fig. \ref{fig:dunnloc}. As observed, the $40\times40$ patch in the top-left corner of the 2D image is found as the best location in terms of Dunn's index for DCT analysis. That shows the lower order frequencies coefficients are related to detection of activities.

In addition, DCT is applied on the original $80\times80$ image so that, 54 DCT coefficients were selected similarly and used for clustering. The results are compared with the proposed local patching strategy. Table \ref{Table:LPE} presents the average testing accuracies of the DCT analysis over the 4-subjects LOOCV for the original image and the selected local patch.

\begin{table}[h!]
\centering
 \caption{Average and standard deviations of  testing accuracies based on K-Means and K-Medoids using DCT features from the raw data $80\times80$ single patch and those from the selected $40\times40$ local patch features over 4 rounds of LOOCV.}
 \scalebox{1.28}{
 \begin{tabular}{|c c a |} 
 \hline
 &  DCT Raw Data & \textbf{DCT-Based Method} \\ [0.5ex] 
 \hline\hline
 K-Means & 63.5\%$\pm$8.64 & 75\%$\pm$5.74  \\ 
 K-Medoids & 62\%$\pm$9.89 & 77\%$\pm$4.58  \\ 
 \hline
\end{tabular}}
\label{Table:LPE}
\end{table}

\subsection{Entropy analysis results}
\begin{figure}[h]
\centering
\includegraphics[scale=0.5]{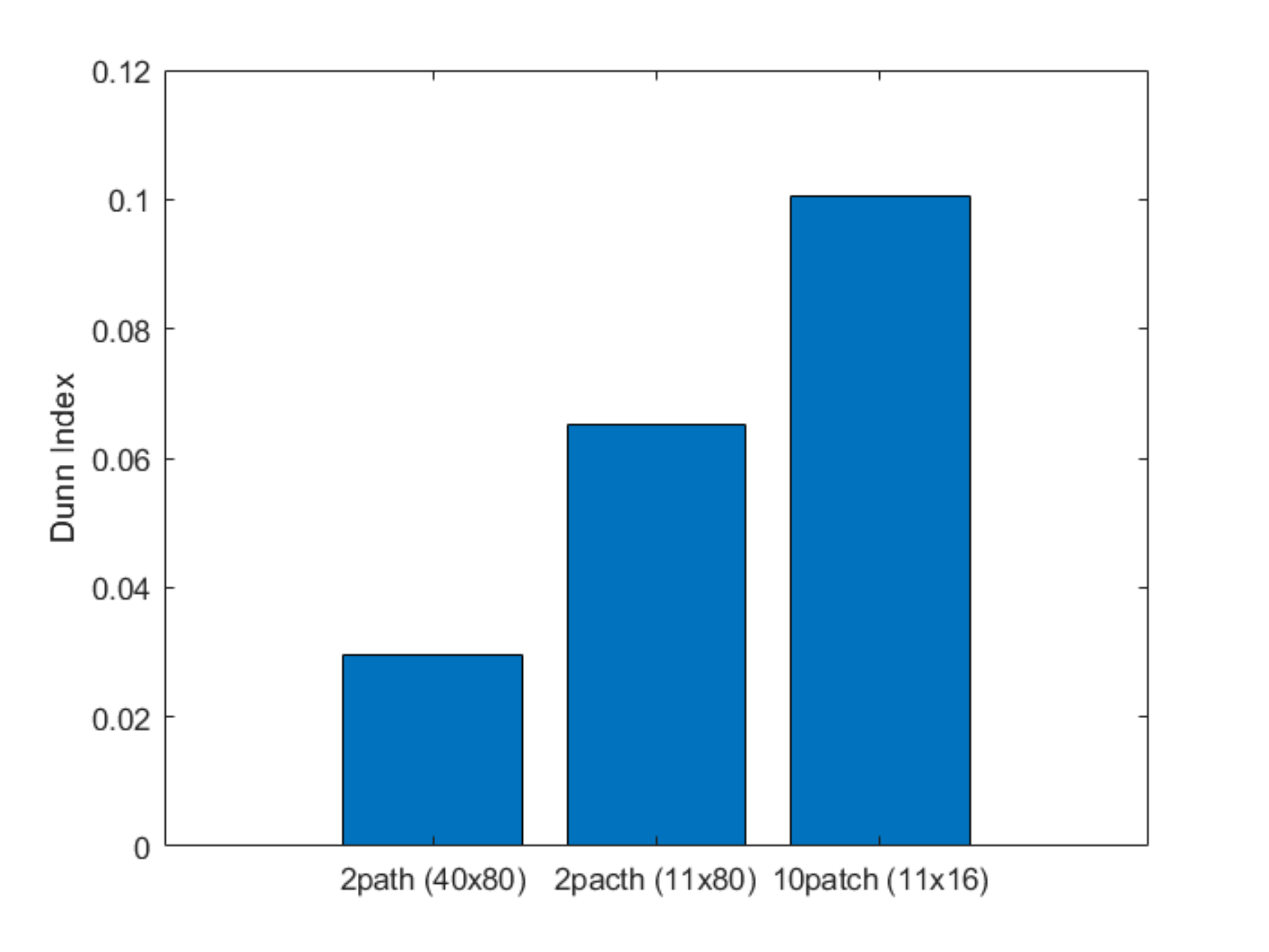}
\caption{Illustration of the Dunn's index for the three patching strategies used for feature extraction based on entropy analysis.}
\label{fig:Entropydunncomp}
\end{figure}

For the entropy analysis, the three patching strategies depicted in Fig. \ref{fig:entropypatch} are considered. The first two strategies resulted into 2-dimensional features, while the last patching strategy resulted into 10-dimensional feature vectors. The results of the Dunn's indexes were computed using the K-Means clustering and presented in Fig. \ref{fig:Entropydunncomp}. As can be seen, the last patching strategy obtained the highest Dunn's index and therefore, it was selected for analysis. This result was expected, because the last strategy considers a higher number (10) of smaller patches. This represents local patterns variations better compared to the other two strategies. The other two strategies consider a fewer number (2) of larger local areas, which leads to poorer entropy computation.


\subsection{CAE and CVAE robustness evaluation}
\begin{figure}[h]
\begin{subfigure}{0.235\textwidth}
\includegraphics[width=\textwidth]{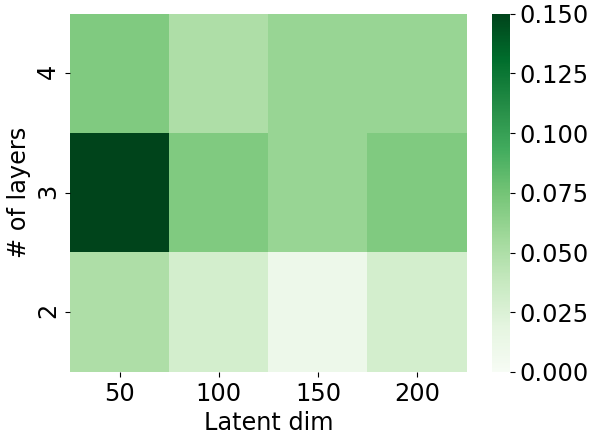}
\caption{}
\label{fig:cae_robust}
\end{subfigure}
\begin{subfigure}{0.235\textwidth}
\includegraphics[width=\textwidth]{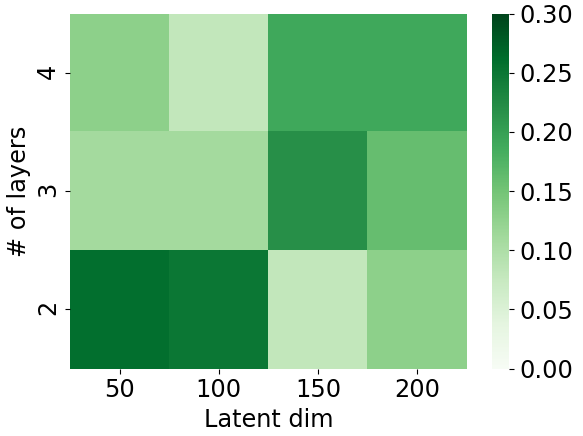}
\caption{}
\label{fig:cvae_robust}
\end{subfigure}
\caption{(a) CAE and (b) CVAE robustness evaluation using Dunn's index.}
\label{fig:cae_cvae}
\end{figure}

The two deep NN architectures CAE and CVAE are evaluated addressing two criteria: 1) the number of hidden layers; and 2) the number of extracted features as latent dimension. In regards to the number of hidden layers, two, three, and four hidden layers are considered. The latent dimension is incorporated with the number of data samples in this study. The considered latent features are 50, 100, 150, and 200. Since the study is unsupervised, the true labels are seen as unknown. Hence, Dunn's index is selected for measuring the wellness of clusters separation. Fig. \ref{fig:cae_cvae} reveals the Dunn's index for each experiment for CAE and CVAE.

Based on the results presented in Fig. \ref{fig:cae_robust}, three hidden layers with 50 extracted features are selected for CAE. In regards to the CVAE architecture, two hidden layers with 50 extracted features are selected based on the results presented in Fig. \ref{fig:cvae_robust}.

\subsection{Comparison of the Average training and testing accuracies for K-Means and K-Medoids using all feature extraction techniques}

The average training and testing accuracies with standard deviations over 4-subjects LOOCV for K-Means and K-Medoids using all feature extraction methods are illustrated in Table \ref{Table:training} and Table \ref{Table:testing}.

As it can be observed, the two superior architectures are the local DCT-based frequency features extracted from the local patches and CVAE encoded features. After that, the local entropy analysis achieved the best results. K-Medoids has better performance than K-Means using the local DCT-based coefficient features. On the other hand, CVAE encoded features are better incorporated with K-Means. The results of the K-Means and K-Medoids are very similar in the case of local entropy-based features. In addition, CAE, PCA and 2DPCA have worse performance, while 2DPCA shows a minor improvement to PCA for K-Medoids.

\begin{figure}[h]
\begin{subfigure}{0.235\textwidth}
\includegraphics[width=\textwidth]{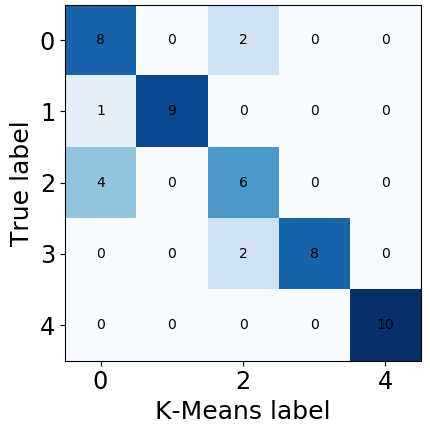}
\caption{}
\label{fig:cm_kmeans_glcm}
\end{subfigure}
\begin{subfigure}{0.235\textwidth}
\includegraphics[width=\textwidth]{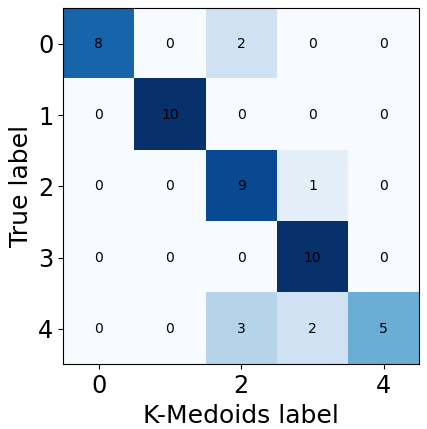}
\caption{}
\label{fig:cm_kmedoids_glcm}
\end{subfigure}
\caption{Confusion matrices for DCT features using K-Means (a) and K-Medoids (b) for different subjects.}
\label{fig:cm_glcm}
\end{figure}

In order to evaluate the two superior architectures' performance for different activity groups, confusion matrices for K-Means and K-Medoids are visualised. The following matrices in Fig. \ref{fig:cm_glcm} consider the proposed local DCT-based method over 4-subjects LOOCV for K-Means (Fig. \ref{fig:cm_kmeans_glcm}) and K-Medoids (Fig. \ref{fig:cm_kmedoids_glcm}).

\begin{figure}[h]
\begin{subfigure}{0.235\textwidth}
\includegraphics[width=\textwidth]{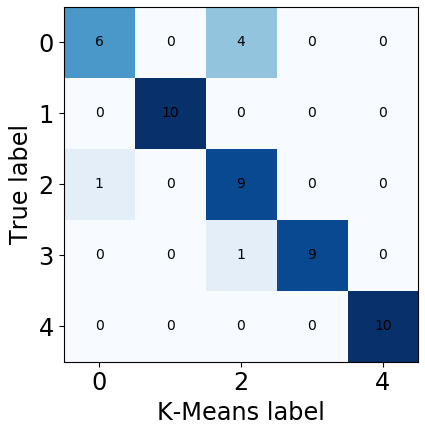}
\caption{}
\label{fig:cm_kmeans}
\end{subfigure}
\begin{subfigure}{0.235\textwidth}
\includegraphics[width=\textwidth]{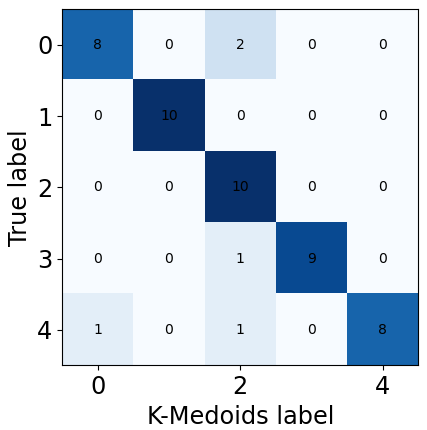}
\caption{}
\label{fig:cm_meds}
\end{subfigure}
\caption{Confusion matrices for CVAE encoded features using K-Means (a) and K-Medoids (b) for different subjects.}
\label{fig:cm}
\end{figure}

The confusion matrices for the CVAE encoded features with K-Means clustering and K-Medoids clustering are visualised in Fig. \ref{fig:cm_kmeans} and Fig. \ref{fig:cm_meds} respectively.

Both feature extraction strategies results show confusion of some activities. Further analysis of the results will be presented in the discussion section.

\begin{table*}[t]
  \caption{Average training accuracies with standard deviations of K-Means and K-Medoids using the five feature extraction methods over 4-subjects LOOCV.}
  \centering
  \scalebox{1.28}{
  \begin{tabular}{| c a a c c c c|} 
 \hline
   & \thead{DCT-Based \\ Method} &  \thead{Entropy-Based \\ Method} & CVAE & CAE & PCA & 2DPCA \\ [0.1ex] 
 \hline\hline
 K-Means & 80\%$\pm$4.41 & 69.5\%$\pm$3.23 & 83.75\%$\pm$3.11 & 52\%$\pm$4.74 & 64.25\%$\pm$7.32 & 67\%$\pm$10.29 \\
 K-Medoids & 77.5\%$\pm$8.87 & 69.5\%$\pm$4.05 & 79.75\%$\pm$3.11 & 54.5\%$\pm$4.55 & 47\%$\pm$4.41 & 64.5\%$\pm$5.93 \\[0.5ex] 
 \hline
  \end{tabular}}
\label{Table:training}
\end{table*}

\begin{table*}[t]
  \caption{Average testing accuracies with standard deviations of K-Means and K-Medoids using the five feature extraction methods over 4-subjects LOOCV.}
  \centering
  \scalebox{1.35}{
  \begin{tabular}{| c a a c c c c|} 
 \hline
   & \thead{DCT-Based \\ Method} &  \thead{Entropy-Based \\ Method} & CVAE  & CAE & PCA & 2DPCA \\ [0.1ex] 
 \hline\hline
 K-Means &  75\%$\pm$5.74 & 72\%$\pm$7.11 & 84\%$\pm$5.09 & 66\%$\pm$8.6 & 58.5\%$\pm$1.65 & 57.5\%$\pm$2.59 \\
 K-Medoids & 77\%$\pm$4.58 & 72\%$\pm$8.48 & 82\%$\pm$4.89 & 65\%$\pm$5.74 & 57\%$\pm$2.23 & 64\%$\pm$8.12 \\[0.5ex] 
 \hline
  \end{tabular}}
\label{Table:testing}
\end{table*}

\subsection{Visualisation results}
\begin{figure*}\centering
\includegraphics[width=1\linewidth]{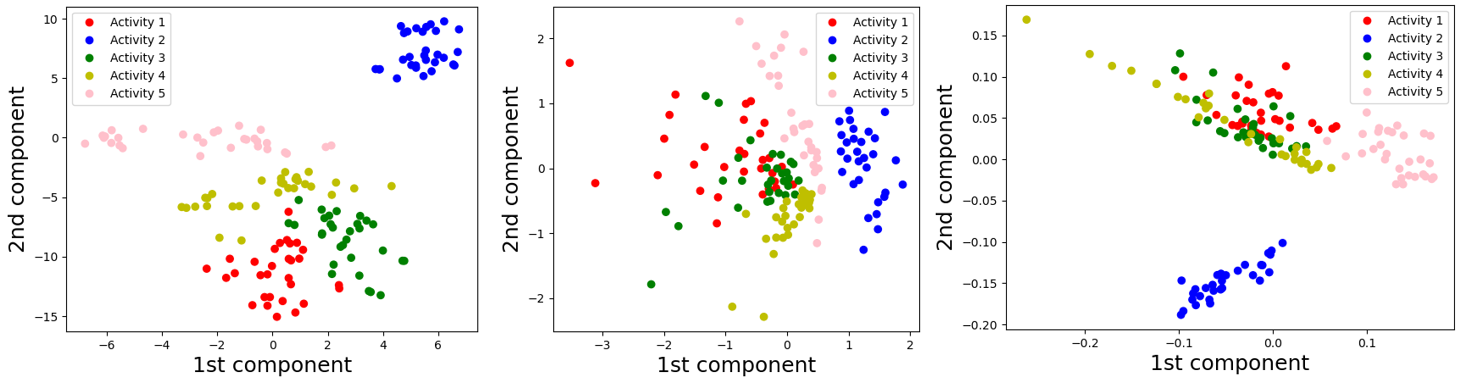}
\caption{Visualisation of the transformed raw data $D_{n \times 6400}$ using t-SNE, MDS, and LLE.}
\label{fig:visual_befor}
\vspace{-1.5em}
\end{figure*}

Visualization of the raw data and CVAE encoded features are performed using t-SNE, MDS and LLE methods. Here, the actual data labels are used to map the samples. Initially, the original data $D_{n \times m}$, where $m=6400$, is transformed and visualised in a 2-dimensional space as shown in Fig. \ref{fig:visual_befor}. As illustrated, t-SNE performs reasonable separability between the classes, while there are some overlapping clusters in the case of MDS and LLE.

\begin{figure*}\centering
\includegraphics[width=1\linewidth]{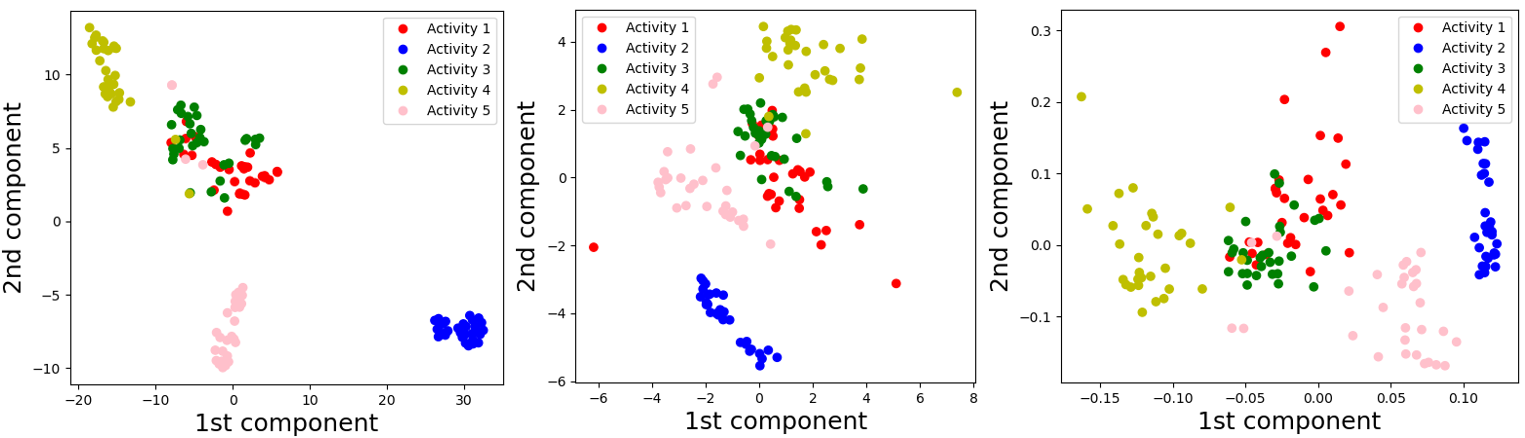}
\caption{Visualisation of the transformed encoded data $D_{n \times 50}$ using t-SNE, MDS, and LLE.}
\label{fig:visual_after}
\vspace{-1.5em}
\end{figure*}

In order to perform comparison, the second scenario of data visualisation is concerned by transforming the encoded data $D_{n \times 50}$ using t-SNE, MDS and LLE to a 2-dimensional space as seen in Fig. \ref{fig:visual_after}. Considering the illustration, all three methods showcase improvements in terms of cluster separability. However, there is still overlapping between the clusters. Since dimension reduction from 50 encoded features into only two features is a significant reduction in the number of features, no accurate clustering is expected using these two dimensional features. This is tested by applying K-Means and K-Medoids on the two t-SNE features. The average testing accuracies are 42\% and 45\% for K-Means and K-Medoids respectively. Hence, these manifold learning methods are good for visualisation, but are not necessarily accurate for clustering.

\section{Discussion}\label{Section:Discussion}
The local patching strategy incorporated with DCT improved the average training and testing accuracies by 10\%-15\% for the two scenarios as seen in Table \ref{Table:LPE}. It has been solely validated using an unsupervised metric for the scope of this study. As such, the method can be applied for other supervised or unsupervised studies with Doppler radar data. The local patching strategies can even be improved when used in a supervised framework, because the average validation accuracies allow optimum estimation of the method's parameters.


The confusion matrices for the architecture local DCT-Based+K-Means and local DCT-Based+K-Medoids in Fig. \ref{fig:cm_glcm} reveal that the activities walking (1) and jumping (3) are problematic as they are frequently confused. Similarly, the CVAE-based architecture in Fig. \ref{fig:cm} shows confusion between walking (1) and jumping (3). As it can be observed from the confusion matrices, the walking (1), jumping (3) and standing (5) classes are seen as problematic. More data can be collected in order to improve the results with a higher number of subjects. Considering the clustering methods, local DCT features are better incorporated with K-Medoids. On the other hand, the CVAE encoded features are more correctly clustered with K-Means. In addition, the results of both clustering strategies were similar in the case of local entropy-based features. That is due to the lower resolution of the entropy features compared to DCT and CVAE techniques. In terms of efficiency, K-Means is shown to execute faster than K-Medoids \cite{KMeans:KMedoids}.


Considering the employed feature extraction architectures, it can be concluded that the encoded data with CVAE is superior in comparison with the proposed local DCT-based and local entropy-based feature extraction methods in terms of accuracy. On the other hand, the local DCT-based and local entropy-based analyses are less complex and more easily implemented. The computational time for local DCT-based method and local entropy-based method is noticeably smaller in comparison with CVAE as seen in Table \ref{Table:time_complexity}. 

\begin{table}[h!]
\centering
 \caption{Total computational time over 600 samples of the four folds for the two proposed feature extraction methods as well as CVAE.}
 \scalebox{1.28}{
 \begin{tabular}{|c a a c |} 
 \hline
 & \thead{DCT-Based \\ Method} &  \thead{Entropy-Based \\ Method} & CVAE \\ [0.5ex] 
 \hline\hline
 Time & 0.58 s & 1.1074 s & 698.39 s  \\ 
 \hline
\end{tabular}}
\label{Table:time_complexity}
\end{table}

The reason is that, the local DCT-based and local entropy-based features are simply derived from the local patches. On the other hand, CVAE encoded features are mainly the results of convolution of the cubic Doppler images of size $2\times100\times32$ with several filter types. Furthermore, the convolutional filters weights of both encoding and decoding structure are learnt based on an optimisation process using the objective function, which is computationally more complex rather than the other proposed strategies. Then, learning the local DCT-based and local entropy-based features from new coming datasets will be faster than the CVAE features. Given the advantages and disadvantages of each strategy, the overall recognition performances of them are reasonable and can be applied to other unsupervised project scenarios. PCA and CAE are not seen as successful due to their limitations. In regards to CAE, the method does not include a regularization term, which is prone to producing inappropriate representations of the data samples in terms of classification. PCA preserves the global structure of the data, but fails to retain local dependencies in the lower dimensional space.

Considering the described individual feature extraction methods, one possible idea is to fuse the features from different strategies. However, there are reasons not to consider that for the current dataset. Given the limited number of samples in this study compared to the high number of features from most techniques, this will increase the dimensionality of the feature space. Therefore, it does not improve the accuracy. That is tested for the fusion of the local DCT-based features and local entropy-based features and no improvement was observed. In addition, since the clustering strategies are based on computation of the features distance and data fusion might require normalization of the heterogeneous feature types, that can also influence the clustering results.

The average training accuracies and average testing accuracies report a slightly bigger standard deviation in some cases. This is explained by the fact that 4-subject LOOCV is applied. As such, in some of the folds, the retained subject data for testing appears very different from the subjects data for training. However, the achieved accuracies are still reasonable for an unsupervised framework. That is a positive sign for the potential use of such strategies for e-Healthcare purposes.

The three manifold learning methods considered in this study can be extremely useful for data visualisation problems. As seen in Fig. \ref{fig:visual_befor}, the t-SNE visualisation is better in comparison with MDS and LLE. Since the unsupervised CVAE-based architecture is known to provide a reasonable separation between the clusters, it is compared against the raw data. The results show better separation for all three visualisation techniques shown in Fig. \ref{fig:visual_after}. It can be concluded that the proposed deep CVAE improves the separability between the classes, which can boost the visualisation results for different manifold learning methods. For high-dimensional data visualisation purposes, the CVAE encoded features can be used prior manifold learning. Despite the successful separation of classes, it can be observed that walking (1) and jumping (3) have overlapping samples. This is also observed in the confusion matrices in Section \ref{SecIV:Results}, where these two activities are commonly mis-classified. Additionally, the samples from running (2) are the most accurately separated from the remaining samples from the other activities. This finding is also evident when the manifold learning methods are applied on the raw data.

The achieved results are comparable with the previous supervised research framework on the same dataset in \cite{Doppler:Healthcare}. In that work, SVD, PCA and physical features are used for feature extraction. The average testing results of all three feature extraction methods combined with SVM for classification was reported to be more than 80\% for different sizes of the training set. Similar result is observed with the proposed unsupervised local DCT-based method, and CVAE, where the average testing accuracies are more than 80\% for the 4-subject LOOCV.
\section{Conclusion}\label{Section:Conclusion}
This work studies the employment of Doppler radar for daily activity recognition using an unsupervised framework. The results of this study push the applications of Doppler radar data in healthcare one step forward to practice by enabling the recognition capability without label or with poor labelling. In particular, the analysis architecture includes unsupervised feature extraction followed by clustering strategies. Four different categories of unsupervised feature extraction, namely, digital image frequency analysis based on DCT, entropy analysis, convolutional filtering based on deep autoencoders architectures CAE and CVAE, and the state of the art PCA and 2DPCA techniques, were employed. More specifically, two unsupervised methods for extraction of local DCT-based and local entropy-based features were proposed. The proposed two local patches-based methods for feature extraction exhibited an improvement of 5\%-10\% average testing accuracies compared to conventional CAE, PCA, and 2DPCA. On the other hand, the CVAE encoded features were superior with average testing accuracies of 84\% and 82\% for K-Means and K-Medoids respectively. Considering the expensive computational time for CVAE, the two proposed local DCT-based and local entropy-based methods provide a reasonable trade-off between time and accuracy. Regarding the unsupervised scenario, that is a positive sign for the potential use of these proposed techniques.

Three manifold learning methods for high-dimensional data visualisation are considered in this study - t-SNE, MDS and LLE. Visualization of the features using these three methods are compared. The results revealed that the clusters have a better separation with all three methods when the visualisation was performed on the CVAE encoded data. Finally, this project can serve as a reasonable application with two proposed unsupervised feature extraction methods and a visualisation framework for project scenarios with poor labeling for both activity clustering and data visualisation.

\section*{Acknowledgment}
Yordanka Karayaneva thanks the sponsorship of the Data Driven Research Innovation (DDRI) at Coventry University, United Kingdom.

\vskip -2\baselineskip plus -1fil
\begin{IEEEbiography}[{\includegraphics[width=1in,height=1.25in,clip,keepaspectratio]{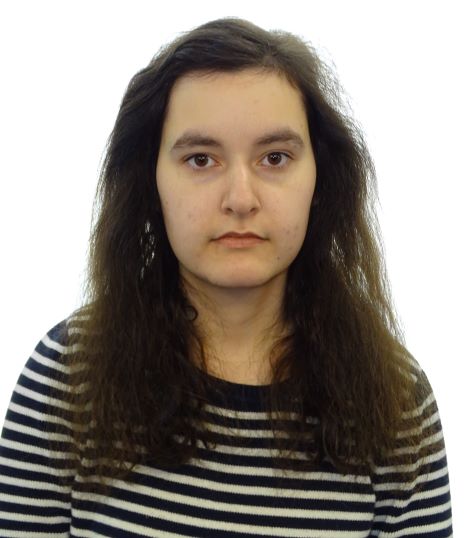}}]{Yordanka Karayaneva} is a PhD student in computer science at Coventry University, UK. She received the B.Sc degree in computer science from Coventry University, UK in 2017.

From June to July 2016, she was a Research Intern at Coventry University in reinforcement learning. Yordanka Karayaneva was again a Research Intern at Coventry University from June to August 2017 in distributed systems. She held the position PhD Teaching Assistant at Coventry University from September 2018 to April 2019. Yordanka Karayaneva has published conference papers in IEEE and ACM venues. Her research interests include machine learning, signal processing, object recognition and computer vision.
\end{IEEEbiography}

\begin{IEEEbiography}[{\includegraphics[width=1in,height=1.25in,clip,keepaspectratio]{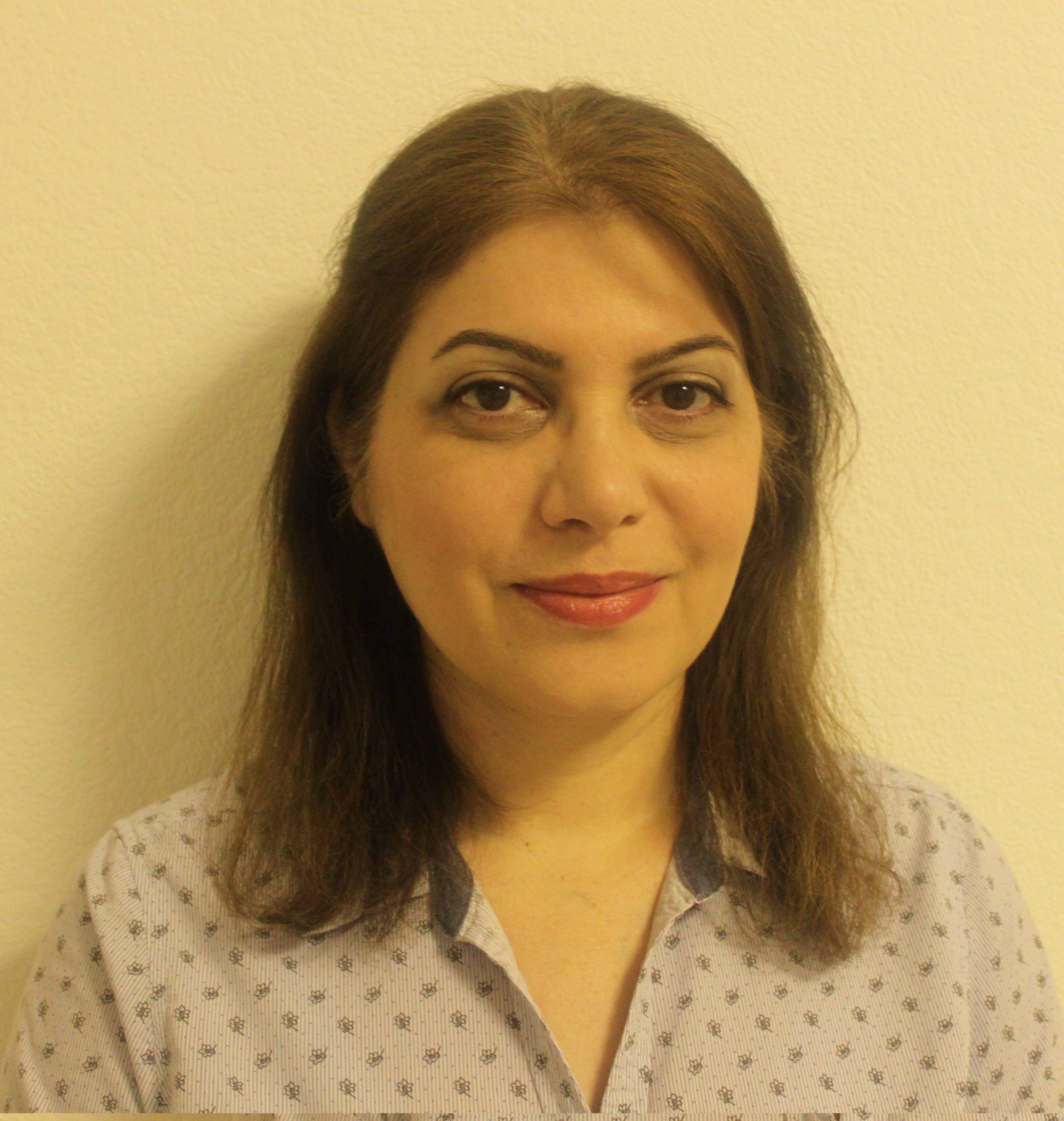}}]{Dr. Sara Sharifzadeh} received her B.Sc and M.Sc degrees in Electronics Engineering from University of Mazandaran, Iran, in  2003 and 2007 respectively. She also received an MSc in Multimedia Technologies from Universitat Autonoma de Barcelona, Spain in 2010. She received her PhD in Computer Science from Technical University of Denmark in 2015.

From 2015 to 2017, she was postdoctoral research associate in data science, at Loughborough University, UK. She is a lecturer at Coventry University since 2018. Her research interests are machine learning, artificial intelligence and their application on analysis of digital signals, images and 3D point clouds. She has published in several leading conferences and journals. Her current research is focused on signal and image analysis for gesture recognition, monitoring, and remote sensing applications.
\end{IEEEbiography}
\vskip -2\baselineskip plus -1fil
\begin{IEEEbiography}[{\includegraphics[width=1in,height=1.25in,clip,keepaspectratio]{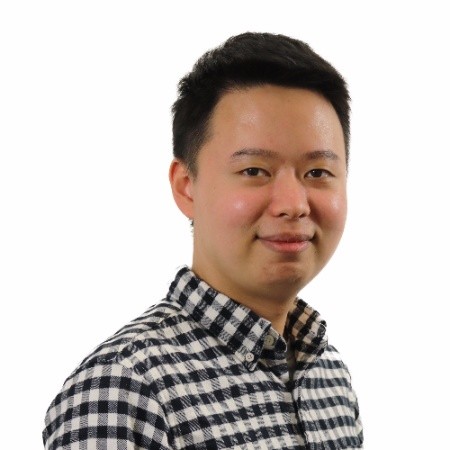}}]{Wenda Li} 
received the M.Eng. and Ph.D. degree from the University of Bristol in 2013 and 2017 respectively. He worked at University of Birmingham as a Research Fellow before joining University College London. He is a Research Fellow in the Department of Security \& Crime Science at University College London. His research focuses on the signal processing for passive radar and high-speed digital system design for wireless sensing applications in healthcare, security and positioning. His research in passive WiFi radar has led to a number of IEEE conference and journal publications. 
\end{IEEEbiography}
\vskip -2\baselineskip plus -1fil
\begin{IEEEbiography}[{\includegraphics[width=1in,height=1.25in,clip,keepaspectratio]{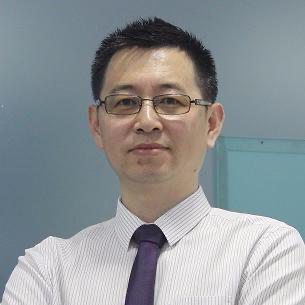}}]{Prof. Yanguo Jing} received the B.Sc and M.Sc degree in Computer Science, Dalian Maritime University, China, in 1997 and 2000 respectively. He received a PhD in Computer Science from Heriot-Watt University, UK, in 2004.

From 2002-2016, he was an Associate Professor at London Metropolitan University, UK. From 2016, he is a Professor at Coventry University, UK. His research is mainly on user modelling, artificial intelligent and machine learning. He is a member of IEEE and IET, a fellow of BCS and CITP. His current work focuses on machine learning methods in intelligent applications. Prof. Jing was a recipient of the Best Paper Award in the 11th International Conference on Developments in eSystems Engineering (DeSE) in 2018.

\end{IEEEbiography}
\vskip -2\baselineskip plus -1fil
\begin{IEEEbiography}[{\includegraphics[width=1in,height=1.25in,clip,keepaspectratio]{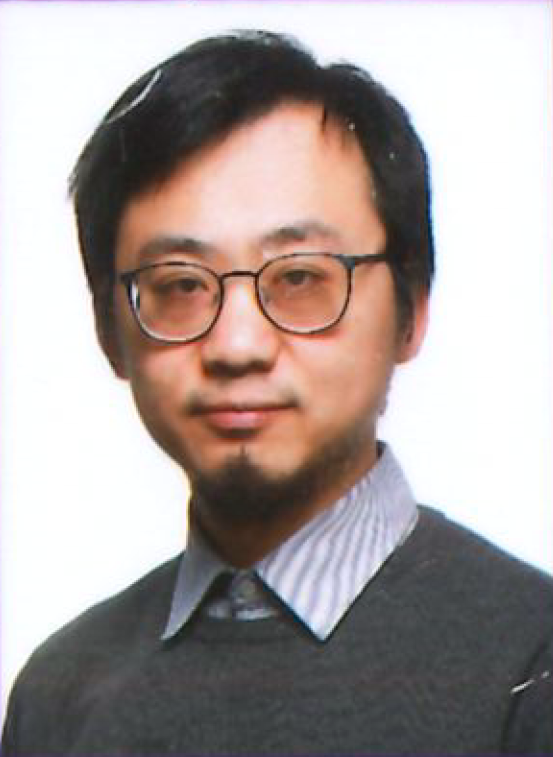}}]{Dr. Bo Tan} received the B.Sc and M.Sc degree in communications engineering, Beijing University of Posts and Telecommunications, in 2004 and 2008 respectively. He received a PhD in Institute for Digital Communications from the University of Edinburgh, UK, in 2013.

From 2012 to 2016, he was postdoc research associate in University College London and University of Bristol. He was a lecturer in Coventry University during 2017 and 2018. From 2019, he is a Tenure Track Assistant Professor at Tampere University, Finland. His research is mainly on radio signal processing in radar and wireless communications systems, also machine learning methods of sensing data. His research has led to 50 academic publications and US patent. He is the member of IEEE and ACM, active reviewer of multiple IEEE and IET journals in sensing and communications. His current work focuses on machine learning enabled wireless sensing and joint radar-communications design in intelligent machines.

\end{IEEEbiography}
\EOD

\end{document}